%% file: template.tex
\documentclass{article}

\usepackage{amsmath} 
\usepackage{arxiv}

\usepackage[utf8]{inputenc} 
\usepackage[T1]{fontenc}    
\usepackage[colorlinks=true, urlcolor=blue]{hyperref}       
\usepackage{url}            
\usepackage{booktabs}       
\usepackage{amsfonts}       
\usepackage{nicefrac}       
\usepackage{microtype}      
\usepackage{multirow}
\usepackage{multicol}
\usepackage{lipsum}
\usepackage{graphicx}
\usepackage{caption}
\usepackage{subfigure}
\graphicspath{{figures/}}
\usepackage{color}
\usepackage{caption}
\usepackage[toc,page]{appendix}
\usepackage{hhline,array}
\makeatother
\usepackage{lineno,amstext}

\author{S. Ece Ada, Emre Ugur, H. Levent Akin\\
  Department of Computer Engineering\\
  Bogazici University\\
  Istanbul,Turkey\\ \{ece.ada, emre.ugur, akin\}@boun.edu.tr}

\begin{document}
\title{Generalization in Transfer Learning}
\maketitle

\begin{abstract}
\input{files/abstract.tex}
\end{abstract}

\section{Introduction}
\input{files/introduction.tex}
\section{Related Work}
\input{files/relatedwork.tex}
\section{Method}
\input{files/method.tex}
\subsection{Background}
\input{files/background.tex}

\subsection{Proposed Method}
\subsubsection{Policy Buffer}
\label{sec:policybuffer}
\input{files/policybuffer.tex}

\subsubsection{Strict Clipping}
\label{sec:strictclipping}
\input{files/strictclipping.tex}

\subsubsection{Regularization in Adversarial Reinforcement Learning}
\label{sec:rarltrl}
\input{files/rarltrl.tex}

\section{Experimental Setup}
\input{files/expsetup.tex}

\section{Results}
\label{chapter:experiments-and-results}
\input{files/expandresults.tex}

\subsection{Humanoid Source Environment}
\label{subsec:humanoidsource}
\input{files/sourcehumanoid.tex}
\subsection{Hopper Source Environment}
\label{subsec:hoppersource}
\input{files/sourcehopper.tex}

\subsection{Regularization via PPO Hyperparameter Tuning}
\label{subsec:regppoexp}
\subsubsection{The Morphology Experiments}
\label{subsubsec:shorttall}
\input{files/shorttall.tex}

\subsubsection{The Friction Environment}
\input{files/fric_exp.tex}

\subsubsection{The Gravity Environments}
\input{files/grav_hum.tex}

\subsection{Regularization in Adversarial Reinforcement Learning}
\label{subsec:regadvexp}
\subsubsection{The Morphology Experiments}
\label{sec:torsomasshopper}
\input{files/torsomasshopper.tex}

\subsubsection{The Gravity Environments}
\label{sec:grav_hopper}
\input{files/grav_hopper.tex}

\section{Conclusion}
\input{files/conclusion.tex}

\bibliographystyle{unsrt}  
\bibliography{references}
\newpage
\begin{appendix}
\section{Hyperparameters}
\label{appendix:hyperparameters}
\setcounter{table}{0}
\begin{table}[!htbp]
	\vskip\baselineskip 
	\caption[Hyperparameters]{Source Task Training Hyperparameters}
	\begin{center}
		\begin{tabular}{|c|c|c|c|c|c|c|c|}\hline
			\textbf{Hyperparameters}&\textbf{Symbol}& \multicolumn{6}{c|}{\textbf{Values}}\\\hline
			Clipping Parameter &$\varepsilon$& 0.01 & 0.025 & 0.05 & 0.1& 0.2& 0.3\\\hline
			Batch size& $b$ &  \multicolumn{3}{c|}{64}& \multicolumn{3}{c|}{512} \\\hline
			Step Size &$\alpha$ & \multicolumn{3}{c|}{0.0001} & \multicolumn{3}{c|}{0.0003}\\\hline
			Curriculum Parameter &$\chi$ & \multicolumn{3}{c|}{0.3} & \multicolumn{3}{c|}{0.5}\\\hline
			Learning Schedule& &  \multicolumn{3}{c|}{constant} & \multicolumn{3}{c|}{linear}\\\hline
			Clipping Schedule& &  \multicolumn{3}{c|}{constant} & \multicolumn{3}{c|}{linear}\\\hline    
			Entropy Coefficient& $\beta_{ pro },\beta_{ adv }$ &  \multicolumn{3}{c|}{0.01}& \multicolumn{3}{c|}{0.03} \\\hline
			Trajectory Size &$H$ & \multicolumn{6}{c|}{2048}\\\hline
			Discount& $\gamma $ &\multicolumn{6}{c|}{0.99}\\\hline
			GAE Parameter& $\lambda $ &\multicolumn{6}{c|}{0.95}\\\hline
			Adam Optimizer & $\beta_{ 1 } $ & \multicolumn{6}{c|}{0.9}\\\hline
			Adam Optimizer & $\beta_{ 2 }$ & \multicolumn{6}{c|}{0.999}\\\hline
			Number of Epochs& & \multicolumn{6}{c|}{10}\\\hline
			Number of Hidden Layers& & \multicolumn{6}{c|}{2}\\\hline
			Hidden Layer Size& & \multicolumn{6}{c|}{64}\\\hline
			Activation Function& & \multicolumn{6}{c|}{tanh}\\\hline
		\end{tabular}
	\end{center}
\end{table} 

\end{appendix}

\end{document}

%% file: files/abstract.tex
Agents trained with deep reinforcement learning algorithms are capable of performing highly complex tasks including locomotion in continuous environments. We investigate transferring the learning acquired in one task to a set of previously unseen tasks. Generalization and overfitting in deep reinforcement learning are not commonly addressed in current transfer learning research. Conducting a comparative analysis without an intermediate regularization step results in underperforming benchmarks and inaccurate algorithm comparisons due to rudimentary assessments. In this study, we propose regularization techniques in deep reinforcement learning for continuous control through the application of sample elimination, early stopping and maximum entropy regularized adversarial learning. First, the importance of the inclusion of training iteration number to the hyperparameters in deep transfer reinforcement learning will be discussed. Because source task performance is not indicative of the generalization capacity of the algorithm, we start by acknowledging the training iteration number as a hyperparameter. In line with this, we introduce an additional step of resorting to earlier snapshots of policy parameters to prevent overfitting to the source task. Then, to generate robust policies, we discard the samples that lead to overfitting via a method we call strict clipping. Furthermore, we increase the generalization capacity in widely used transfer learning benchmarks by using maximum entropy regularization, different critic methods, and curriculum learning in an adversarial setup. Subsequently, we propose maximum entropy adversarial reinforcement learning to increase the domain randomization. Finally, we evaluate the robustness of these methods on simulated robots in target environments where the morphology of the robot, gravity, and tangential friction coefficient of the environment are altered.

%% file: files/introduction.tex
Transfer learning refers to means of utilizing knowledge gained in one or multiple tasks for unseen novel tasks \cite{lazaric2012transfer}. In transfer learning, the source task is the origin of knowledge whereas the target tasks are the test tasks that are not seen during training. A robot is expected to be robust to an unbounded number of tasks in real life. For instance, a bipedal robot will be required to be robust to different ground frictions, gravity, inclinations, and abrupt winds. Training a robot for all possible scenarios is time-consuming, impractical, and expensive. To develop intelligent systems, we expect the robots to generalize the learned model to the novel target task different from the source task, adequately and quickly.

Increasing generalization between tasks via Deep Reinforcement Learning (RL) is a useful strategy to acquire a robust robotic skill set. A common way of assessing generalization is evaluating a deep RL algorithm in a different task. Yet, deep RL algorithms, including the policy gradient methods \cite{schulman2015trust,schulman2017proximal,haarnoja2018soft}, were commonly evaluated by the training performance because test and training task distinction do not exist. After successful results were obtained with policy gradient algorithms like Trust Region Policy Optimization (TRPO) \cite{schulman2015trust}, Proximal Policy Optimization (PPO) \cite{schulman2017proximal} and Soft Actor-Critic (SAC) \cite{haarnoja2018soft}, many efforts have been made to solve the problem of transferring the information from a given task to a different task in RL \cite{finn2017model,frans2017meta,pinto2017robust}.

Quantifying generalization in state-of-the-art deep RL is challenging because the performance of deep RL algorithms is highly dependent on the hyperparameter and framework selection. This nature of deep RL algorithms affects the evaluation of transfer RL algorithms because transfer RL algorithms use deep RL algorithms as baselines. Henderson et al. \cite{henderson2018deep} showed that different hyperparameters yield different results in non-transfer RL tasks. Statistically significant results can be obtained just by tuning the hyperparameters of the baseline algorithms. Yet, some hyperparameters (such as the clipping hyperparameter of PPO that controls the lower bound on the policy gradient loss) were set to fixed values while evaluating the results in many studies \cite{fujita2018clipped,han2019dimension,bansal2017emergent,journals/corr/abs-1710-03641,huang2018reinforcement}. Similarly in the transfer RL domain, the trade-off between the target and the source task performance is oftentimes neglected when determining the hyperparameters. For instance, the best policy in the source task is not necessarily the best and the most robust policy in the target task. 

One way of increasing the robustness of the policy is regularization. Hence, we propose regularization techniques that increase target task performance. We experimentally prove that the failure to acknowledge overfitting as well as diverse dimensionality of the robots induce inaccurate comparisons. On account of this, we first show that source task performance is not indicative of generalization capacity and target task performance. Without regularization of the baselines and the proposed methods, we can not distinguish improvements because the baselines will underperform. In this paper, we propose a method namely \textit{Strict Clipping PPO} (SC-PPO) to discard the samples that cause overfitting by increasing the lower bound on the policy gradient loss. We apply SC by decreasing the clipping parameter in PPO to extremely small values and prove its impact numerically on a high dimensional Humanoid robot and a Hopper robot in the MuJoCo environment \cite{todorov2012mujoco}.  This regularization technique independently increases the performance of the policy in the target task and constitutes a competitive benchmark. We evaluated our method in Humanoid and Hopper benchmarks proposed in \cite{henderson2017benchmark} and \cite{pinto2017robust}. Subsequently, we verified our methods in both novel target tasks that involve transfer to robots with different morphologies and target tasks with an increased range of environmental dynamics. These novel tasks include tall, short, and delivery humanoid target tasks, each having a different center of mass and morphologies than the standard humanoid source task. 

In transfer RL problems where the environmental dynamics of the target task diverges more from the source task, we applied early stopping. We have observed that the policy trained in the source task started to overfit the target task after several policy iterations. In this regard, we attest that earlier iterations of policies perform well in comparative analysis because stopping the training earlier avoids overfitting to the source task. We additionally show that the recognition of the policy iteration number as a hyperparameter increases the performance of state-of-the-art algorithms like PPO and Robust Adversarial Reinforcement Learning (RARL) \cite{pinto2017robust}. 

Regularization techniques can also be used in combination. We perform comparative analysis with RARL to show the generalization capacity of using the combination of SC-PPO and early stopping regularization techniques. We compare RARL algorithm with our methods on the torso mass \cite{pinto2017robust} and gravity \cite{henderson2017benchmark}  benchmarks where the hopper is expected to hop without falling. Table \ref{table:comparisonhopperacrarl} shows the torso mass target tasks generated in the range [1,9] by modifying the mass of the torso and gravity target tasks generated in the range [0.5G,1.75G] by modifying the gravity of the environment. We use the same source task for training where the hopper torso mass is approximately 3.53 units and the gravity is $G=9.81$. Early stopping is used in all of our methods as regularization. Our methods can successfully extrapolate to the target tasks in the corresponding range exceeding RARL in all benchmarks. 

We compare the generalization capacities of different adversarial training methods in the target tasks, namely, entropy bonuses, advantage estimation techniques, and different curriculum learning \cite{shioya2018extending} with RARL. We integrate an entropy bonus in adversarial RL in Maximum Entropy Robust Adversarial Reinforcement Learning (ME-RARL) to increase domain randomization. Our results show that entropy regularized adversarial RL significantly increases generalization in the target tasks. Furthermore, we increase generalization using the advantage estimation component by involving both value function estimator critic networks at each optimization iteration in Section \ref{sec:rarltrl} named Average Consecutive Critic Robust Adversarial Reinforcement Learning (ACC-RARL). Finally, we provide numerous experimental results to prove that our propositions outperform the existing methods significantly and increase the range of the commonly used benchmarks. We hope these techniques will aid in developing competitive benchmarks for future transfer RL research. 

\begin{table*}[thbp]
	\vskip\baselineskip 
	\caption[Average Reward per Episode of the policy trained with ACC-RARL]{Target task extrapolation success range}
	\begin{center}
	\begin{tabular}{|c|c|c|c|}\hline
			\textbf{Task}& \textbf{SC-PPO (ours)}& \textbf{ACC-RARL (ours)}& \textbf{ME-RARL(ours)}\\\hline
			Hopper-v2 Torso Mass &[1,6]&[1,7] &[1,9]\\\hline    
			Hopper-v2 Gravity& [1G,1.5G]& [0.5G,1.75G]&[0.5G,1.75G]\\\hline
		\end{tabular}
	\label{table:comparisonhopperacrarl}
	\end{center}
\end{table*}

In Section II, studies on the transfer RL and adversarial learning are provided. The proposed methods along with the background will be detailed in Section III. Experimental setup and results will be given in Section IV and V respectively. Finally, in Section VI, the conclusion section will include a summary of our contributions, discussion of the results, and future directions for research. 

%% file: files/relatedwork.tex
\subsection{Transfer Reinforcement Learning}
Transfer learning methods are categorized into three branches \cite{deeprlcourse-fa17}. Transferring the knowledge and representation from one task to another task is called \textit{forward transfer learning} \cite{deeprlcourse-fa17}. In a \textit{multi-task transfer learning} setting, the transferred knowledge, and the representation are based on multiple different tasks \cite{tobin2017domain}. Muti-task RL is evaluated by the performance of the agent in multiple target environments. \textit{Meta-learning} on the other hand focuses on reusing the learning experience \cite{finn2017model}. Although the settings in multi-task transfer and meta-learning are the same, their main difference resides in the transferred representation. Most transfer learning problems are subject to the risk of negative transfer. Negative transfer occurs when the algorithm performs worse with transfer than not using any transfer at all \cite{wang2019characterizing}. In this case, initializing the algorithm without any information yields better performance than starting from a suboptimal point that leads to insufficient exploration. Most applications in transfer RL include the transfer of the policy parameters to the new task while avoiding negative transfer. The purpose of these applications is to increase the robustness in the target task via the manipulation of the training phase.

One way of extending the capabilities of existing robotic skills in transfer RL is by increasing only the generalization capacity of the RL component. Because the source and target tasks are identical in deep RL, algorithms' robustness to target tasks generated with domain shift is oftentimes overlooked \cite{zhang2018study}. Zhao \textit{et al.} \cite{zhao2019investigating} studied generalization by parametrizing domain shift where the target task parameters are selected from a different distribution than the source task. In their work, the domain shift was realized by using systematic shifts of the environmental parameters, and noise scales injected to the transition, observation, actuator functions of the RL task.

Regularization is challenging in transfer RL because the performance on the source task is insufficient in determining the performance on the target task. Using unregularized deep RL algorithms results in inaccurate comparative analysis. In this respect, Cobbe \textit{et al.} \cite{cobbe2018quantifying} developed a benchmark for generalization named CoinRun where a high number of tests and training levels could be generated. The generalization capacities of various agents were compared using the ratio of successfully solved target task levels by each agent. Few-shot learning is learning from a few labeled target task data. In zero-shot learning, however, no labeled target task data is provided during test time. Zero-shot performance can be improved using domain randomization \cite{tobin2017domain} where task parameters are randomized when forming the training data. Dropout, L2 regularization, data augmentation, batch normalization, and increasing the stochasticity of the policy and the environment were the regularization techniques used in \cite{cobbe2018quantifying} to increase the zero-shot performance. Still, few shot-learning in the target task was required to tune the hyperparameters for generalization. Similar to \cite{cobbe2018quantifying}, Zhao \textit{et al.} \cite{zhao2019investigating} compared regularization techniques for deep RL such as policy network size reduction, soft actor-critic entropy regularizer \cite{haarnoja2018soft}, multi-domain learning \cite{tobin2017domain}, structured control net \cite{srouji2018structured} and adversarially robust policy learning  (APRL) \cite{mandlekar2017adversarially}. 

\subsection{Adversarial Learning}

Adversarial algorithms are dynamic ways of generating challenging tasks for the agent at each iteration to increase robustness in unseen target tasks \cite{journals/corr/abs-1710-03641}. 
Adversarial learning is among the widely used techniques to increase generalization capacity in robotics \cite{bousmalis2018using,tzeng2015adapting}. In the work of \cite{bousmalis2018using,tzeng2015adapting}, adversarial techniques are successfully utilized using two networks that learn together and advance by competing with each other. More specifically, the discriminator network is optimized to discriminate the real-world image data from the simulation whereas the generator network is optimized to generate simulator images that can fool the discriminator. Similarly in RARL\cite{pinto2017robust}, a separate adversarial network is optimized to destabilize the agent during training. 

Inspired by the efficacious strategy in RARL \cite{pinto2017robust}, Shioya \textit{et al.} \cite{shioya2018extending} proposed two extensions to RARL by varying the adversary policies. First method they proposed was to add a penalty term to the adversary policy's reward function by sampling from the target task to adapt to the transition function of the source task. This is, however, tailored robustness for each target task at hand which requires sampling from the target task similar to the Bayesian update used in Ensemble Policy Optimization (EPOpt) \cite{rajeswaran2016epopt}. The second extension was inspired by curriculum learning \cite{bengio2009curriculum} that selects the adversarial agents based on the progress of learning instead of naively taking the latest adversarial policy. Training the protagonist in harder tasks does not guarantee a more robust policy. The level of difficulty required for high performance on the target task is highly dependent on the transfer problem. Bansal \textit{et al.} \cite{bansal2017emergent} used uniform distribution to determine opponent humanoid policy's iteration from the subset of the latest iterations. In both Shioya \textit{et al.}'s \cite{shioya2018extending} and Bansal \textit{et al.}'s \cite{bansal2017emergent} experiments, using the policy of the latest and the hardest adversary hindered the learning progress of the agent. In \cite{shioya2018extending}, they used multiple adversaries and ranked each sample's performance to determine the set of samples that should be used for optimization. Each adversary policy was maximized using the negative reward of the agent and the sum of KL Divergence from all the other adversary policies to encourage diversity between the adversaries. Experiments were done using Hopper and Walker2d Open AI gym robot control environments\cite{1606.01540} in MuJoCo \cite{todorov2012mujoco} to compare the results to the RARL Algorithm \cite{shioya2018extending}. In both of the environments, using the less rewarding rollouts performed worst. As a contrast, optimizing over the worst performing samples generated more robust policies in Hopper task when tested with different torso masses in EPOpt \cite{rajeswaran2016epopt}.

Adversarial algorithms are generating considerable interest in RL \cite{pinto2017robust,bansal2017emergent}. Our method improves on previous work by employing regularization techniques to succeed in target tasks with higher domain shift. In the comparative analysis, we show that compared to RARL and PPO, the agent trained using our methods can successfully transfer skills to a wider range of unseen target tasks.

%% file: files/method.tex
In this paper, we propose a framework to obtain generalizable policies. Section \ref{sec:policybuffer} describes the Policy Buffer that is used to observe, and store the policies to select the generalizable ones for target task transfer. Section \ref{sec:strictclipping} provides the Strict Clipping component that prevents policy updates when the training samples improve the source task objective above a threshold. A low threshold stabilizes training and regularizes the source task objective. Regularization is applied using different methods in an adversarial framework in Section \ref{sec:rarltrl}. More specifically, the agent is encouraged to explore the environment, different deep RL architectures are applied, and incremental learning is used to decrease overfitting. Before providing the details of our components, we will explain the background, namely the adversarial and deep reinforcement algorithms in the next section.

%% file: files/background.tex
We build our method on top of RARL algorithm which is characterized as a two-player zero sum discounted game \cite{rllab-adv,gym-adv,rllab}. The return of the agent is formalized as follows. 

\begin{equation}
	\begin{aligned}
		R ^ { 1 } &= E _ { s _ { 0 } \sim \rho , a ^ { 1 } \sim \mu( s ) , a ^ { 2 } \sim \nu ( s ) } \left[ \sum _ { t = 0 } ^ { T - 1 } r^ { 1 } \left( s_t , a_t ^ { 1 } , a_t ^ { 2 } \right) \right] 
	\end{aligned}
	\label{eq:rewardrarl}
\end{equation}

Actions $a^1$ are sampled from the policy $\mu$ of the agent and actions $a^2$ are sampled from the policy  $\nu$ of the adversary. $s _ { 0 }$ is the initial state sampled from the initial state distribution $ \rho$. $s_t$ is the state at timestep t and r corresponds to the reward function. The agent maximizes its return whereas the adversary minimizes the return of the agent. Thus, the return of the adversary is $R ^ { 2 }=-R ^ { 1 }$.

We use PPO to update actor-critic networks of the agent and the adversary in RARL experiments as in \cite{rarlbaselines}. Actor network determines the actions the agent takes whereas the critic network estimates the value function used for the advantage function estimation \cite{schulman2015high}. Advantage function estimates how rewarding it is to take the action in the state compared to the value of that state. We use a particular loss function, namely Clipped PPO Loss, $L ^ { C L I P }$ \cite{schulman2017proximal} that optimizes the parameters $\theta$ of the actor policy network as follows.

\begin{equation}
\begin{array} { l l }{L ^ { C L I P } ( \theta )}&{= \hat { \mathbb { E } } _ { t } \left[ \min \left(\frac{\pi_{\theta}\left(a_{t} \mid s_{t}\right)}{\pi_{\theta_{\text {old }}}\left(a_{t} \mid s_{t}\right)} \hat { A } _ { t } , \operatorname { clip } \left(\frac{\pi_{\theta}\left(a_{t} \mid s_{t}\right)}{\pi_{\theta_{\text {old }}}\left(a_{t} \mid s_{t}\right)}, 1 - \varepsilon , 1 + \varepsilon \right) \hat { A } _ { t } \right) \right]}\end{array}
\label{eq:clipCPI}
\end{equation}

Generalized Advantage Estimator (GAE) \cite{schulman2015high} $\hat { A } _ { t }$ is used to optimize the actor network. If the advantage  $\hat { A } _ { t }$ is negative then the probability ratios below $1 - \varepsilon$ are clipped and if the advantage is positive, then the probability ratios above $1 + \varepsilon$ are clipped. Gradient flow does not occur and the samples are discarded if the probability ratio $\frac{\pi_{\theta}\left(a_{t} \mid s_{t}\right)}{\pi_{\theta_{\text {old }}}\left(a_{t} \mid s_{t}\right)}$ is clipped and the expression $\left(\operatorname { clip } \left(\frac{\pi_{\theta}\left(a_{t} \mid s_{t}\right)}{\pi_{\theta_{\text {old }}}\left(a_{t} \mid s_{t}\right)}, 1 - \varepsilon , 1 + \varepsilon \right) \hat { A } _ { t } \right)$ is minimum. $\pi_{\theta}\left(a_{t} \mid s_{t}\right)$ is the current policy being optimized whereas $\pi_{\theta_{\text {old }}}\left(a_{t} \mid s_{t}\right)$ is the previous policy. $\varepsilon$ is the clipping parameter that controls the lower bound on the Clipped PPO Loss.

%% file: files/policybuffer.tex
Policies trained with different hyperparameters show different control patterns. We propose a policy buffer to store these policies that are trained with the same loss function but with different hyperparameters. We add the iteration number to the hyperparameter set for transfer RL benchmarks. In simulated experiments, we show that it is possible to extract a comprehensive set of policies capable of exhibiting various skills from a single source task. 

Our method is inspired by the early stopping regularization technique that is frequently used in supervised learning. Similar to supervised learning, in the transfer RL setting, snapshots of policy parameters taken at different steps have different performances in the target task. We train policies with different hyperparameters and take snapshots of the corresponding policies at each optimization iteration at predetermined intervals and save them in the policy buffer. Previous works trained single policies for a constant number of iterations in the source task. In contrast, we evaluate multiple policies from the buffer to determine the iteration number where the model starts to overfit. 

Our aim here is to generate successful policies for a range of target tasks that are represented by task parameters such as gravity, friction, mass, and center of mass of the robot. Overfitting is predominant in cases where the target tasks deviate significantly from the source task. Inspired by the validation dataset idea used in early stopping regularization for supervised learning, we propose designing a \textit{proxy validation tasks set}. Given source and target task parameters and the policies learned in the source task, we form \textit{proxy validation tasks} between source and target tasks with parameters closer to the target task. Expected rewards of the policies trained with different hyperparameters are informative in finding the generalizable policies from the buffer.

%% file: files/strictclipping.tex
To train the aforementioned policies, we use policy gradient algorithms. In literature, hyperparameter tuning of the RL component is commonly overlooked and a fixed set of hyperparameters is used when integrating the RL component into the transfer domain. More specifically, \textit{Open AI Baselines} framework and most of the literature has been using the clip parameter of 0.2 for continuous control tasks \cite{baselines,schulman2017proximal,bansal2017emergent,journals/corr/abs-1710-03641,huang2018reinforcement,henderson2018deep}.

Gradient clipping in Equation \ref{eq:clipCPI} is generally used to discourage catastrophic displacement in the parameter space. In our work, we additionally propose that it can be used for regularization in transfer RL. In particular, we propose a new regularization technique for PPO, namely \textit{Strict Clipping} (SC) to avoid overfitting by constraining the gradient updates. During training, SC-PPO allows more source task samples to be discarded which would otherwise lead to overfitting. SC-PPO is used to further decrease the gradient movement in the policy parameter space in favor of generalization. This is achieved by decreasing the clipping parameter used in PPO by one or less order of magnitude. We prove that this method is superior to the unregularized RARL baseline in multiple transfer learning benchmarks. 

%% file: files/rarltrl.tex
Introducing an adversary to destabilize the agent using multidimensional simulated forces have generated successful results in continuous control tasks in RARL \cite{pinto2017robust}. However, we experimentally prove that unregularized RARL performs worse than regularized PPO. In contrast to prior work, we acknowledge policy iteration as a hyperparameter in all our comparisons. Thus, we compare our methods in an adversarial framework by forming a policy buffer and extracting the most generalizable policies to increase the robustness of the algorithm. More specifically, we regularized critic networks, increased exploration, and integrated curriculum learning in an adversarial RL framework.

\paragraph{Average Consecutive Critic Robust Adversarial Reinforcement Learning (ACC-RARL) (I)} Similar to actor networks, critic networks are function approximators that are prone to overfitting. We propose Average Consecutive Critic Robust Adversarial Reinforcement Learning (ACC-RARL), which computes advantage estimates using the mean of the critic outputs. Critics are optimized with their actor pair consecutively. We aim to decrease overfitting by using double critic networks with different random initializations and reuse the previous model by including the output of the previously updated critic in the advantage estimation. The temporal difference residual of the approximate value function with discount $\gamma$ at timestep $t$ corresponds to $\delta _ { t }$. The temporal difference residuals of the adversary and the protagonist in ACC-RARL are shown as follows.

\begin{equation}
\begin{aligned}
{\delta _ { t } ^ {protagonist} }  &=(- V_{protagonist} \left( s _ { t } \right)+ V_{adversary} \left( s _ { t } \right))/2 + r _ { t } + \gamma ( V_{protagonist} \left( s _ { t + 1} \right)- V_{adversary} \left( s _ { t+1 } \right))/2\\
{\delta _ { t } ^ {adversary} } &= -{\delta _ { t } ^ {protagonist} } 
\end{aligned}
\label{eq:rarladv}
\end{equation}

\paragraph{Maximum Entropy Robust Adversarial Reinforcement Learning (II)} Entropy bonus is used for exploration by rewarding the variance in the distribution of the action probabilities. Following the entropy bonus idea in \cite{mnih2016asynchronous,cobbe2018quantifying}, we incorporate an entropy bonus $H[\pi^{pro}]$ for the protagonist and an entropy bonus $H[\pi^{adv}]$ for the adversary in the reward functions. Updated reward functions of the adversary $R_{t}(\pi^{adv}_{\theta})$ and the protagonist $R_{t}(\pi^{pro}_{\theta})$ are given

\begin{equation}
\begin{aligned}
R_{t}(\pi^{adv}_{\theta})&= \hat { \mathbb { E } } _ { t }[ R_{t}^{actor}(\pi^{adv}_{\theta}) - R_{t}^{critic}(\pi^{adv}_{\theta}) + \beta_{adv}{H[\pi^{adv}]{\theta}(s_t)}]\\
R_{t}(\pi^{pro}_{\theta})&= \hat { \mathbb { E } } _ { t }[ R_{t}^{pro}(\pi^{pro}_{\theta}) - R_{t}^{critic}(\pi^{pro}_{\theta}) + \beta_{pro}{H[\pi^{pro}]{\theta}(s_t)}]
\end{aligned}
\label{eq:merarl}
\end{equation}

where $\pi^{adv}_{\theta}$, $\pi^{pro}_{\theta}$ and $\beta$ correspond to the policy of the adversary, policy of the protagonist, and the entropy coefficient respectively. Increasing exploration at the cost of decreasing source task performance increases the generalization capacity of the algorithm. Our hypothesis is motivated by the increase in robustness in competitive and adversarial environments \cite{bansal2017emergent,shioya2018extending}. However, we acknowledge that adjusting the stochasticity of the environment while avoiding detrimental impact on learning is challenging. Adding an entropy bonus to the loss function of the adversary increases the domain randomization, affecting the performance of both the protagonist and the adversary. We compare the ME-RARL algorithms: entropy regularized RARL (ERARL) and entropy regularized ACC-RARL (EACC-RARL) in more challenging hopper morphology and gravity benchmarks where SC-PPO is not sufficient. 

\paragraph{Curriculum Learning (III)} Curriculum Learning, focuses on discovering the optimal arrangement of the source tasks to perform better on the target task. Similar to \cite{bansal2017emergent,shioya2018extending}, we utilize curriculum learning by randomizing the adversary policy iterations from different sets of advancement. We compare the target task performance of protagonist policies trained with adversaries randomly chosen from the last predefined number of iterations. We use uniform sampling from the policy set of the adversary \cite{shioya2018extending}. At each iteration, based on the sampling from the $\text{Uniform} ( \chi v , v ) $ distribution \cite{bansal2017emergent}, the adversary from the policy storage is used. Adversaries loaded from and recorded to the buffer during curriculum training become less capable and more inconsistent as the training progress and $\chi$ decreases. In consequence, we intend to show how a variety of design decisions during training affects the target task performance in the pursuit of developing novel regularization techniques and more reliable benchmarks. 

%% file: files/expsetup.tex
Robots are expected to generalize to the target tasks using experience from the source tasks. We use a set of transfer learning experiments in the \textit{Humanoid-v2} and  \textit{Hopper-v2} MuJoCo simulation environments \cite{todorov2012mujoco}. To demonstrate the generalization capability of our methods, we create target tasks through the modification of the environment dynamics \cite{pinto2017robust}. We use hopper environments to compare our methods to RARL \cite{pinto2017robust} and humanoid environments to verify the transfer to tasks where the morphology as well as the functionality of the robot is altered. We use \textit{OpenAI Baselines} framework \cite{baselines} in all our experiments.

The humanoid target tasks are generated to show the performance of the policies trained with SC-PPO and early stopping. Rajeswaran \textit{et al.} altered ground friction to create a target task for a hopper robot \cite{rajeswaran2016epopt}. Similarly, we transfer the learning experience from a source task with ground tangential friction coefficient of 1 to a target task with ground tangential friction coefficient of 3.5. Transferring among morphologically different robots with different limb sizes or torso masses have been a popular multi-task learning benchmark \cite{henderson2017benchmark,rajeswaran2016epopt,pinto2017robust}. Accordingly, we generate three novel target environments: a taller and heavier humanoid robot, a shorter and lighter humanoid robot and a delivery robot that carries a heavy box. 

Hopper environments are used to compare adversarial methods with RARL. Similar to the multi-task transfer learning experiments in \cite{henderson2017benchmark}, we generate target tasks by altering the torso mass of the robot and the gravity of the environment. In \cite{henderson2018optiongan}, 4 target tasks were created by modifying the gravity parameters of the environment in the range [0.50G, 1.50G] where G=-9.81. In our experiments, we use a larger range [0.50G, 1.75G] to discover the range of tasks our method can solve. Similarly for the torso mass experiments, we create target tasks by significantly increasing the range of the torso mass of the robot to [1.0, 9.0] from the range [2.75, 4.5] used in RARL.

%% file: files/expandresults.tex
After detailing source task training implementations in Sections \ref{subsec:humanoidsource} and \ref{subsec:hoppersource}, we provide results of our methods SC-PPO in Section \ref{subsec:regppoexp} and  ACC-RARL, maximum entropy RARL in Section \ref{subsec:regadvexp} using various target tasks. The hyperparameters used in source task training are given in Table \ref{appendix:hyperparameters}. We report on the expected mean and standard deviation of the rewards obtained from 32 randomly seeded identical environments for each target task.

%% file: files/sourcehumanoid.tex
The reward function of the \textit{Humanoid-v2} environment in \cite{1606.01540} consists of a linear forward velocity reward, a quadratic impact cost with lower bound 10, a quadratic control cost, and an alive bonus $C _ {bonus}$.

\begin{equation}
r _ { humanoid } ( s , a ) = 0.25*r _ {v _ { f w d }} +min(5 \cdot 10 ^ { - 7 } * c _ {impact}( s , a ) ,10)+0.1*c _ {control} ( s , a ) + C _ {bonus} 
\label{eq:HumanoidRwdFunction}
\end{equation}

If the \textit{z-coordinate} of the center of mass of the agent is not in the interval [1,2], the episode terminates. The total loss function of the PPO algorithm used to update actor and critic networks is

\begin{equation}
{L _ { t } ^ { C L I P + V F } ( \theta ) } {= \hat { E } _ { t } \left[ L _ { t } ^ { C L I P } ( \theta ) - c _ { 1 } L _ { t } ^ { V F } ( \theta ) \right] }
\label{eq:FinalPPO}
\end{equation}

\begin{figure}[!htbp]
	\centering
	\subfigure[]{
		\includegraphics[width=.317\textwidth]{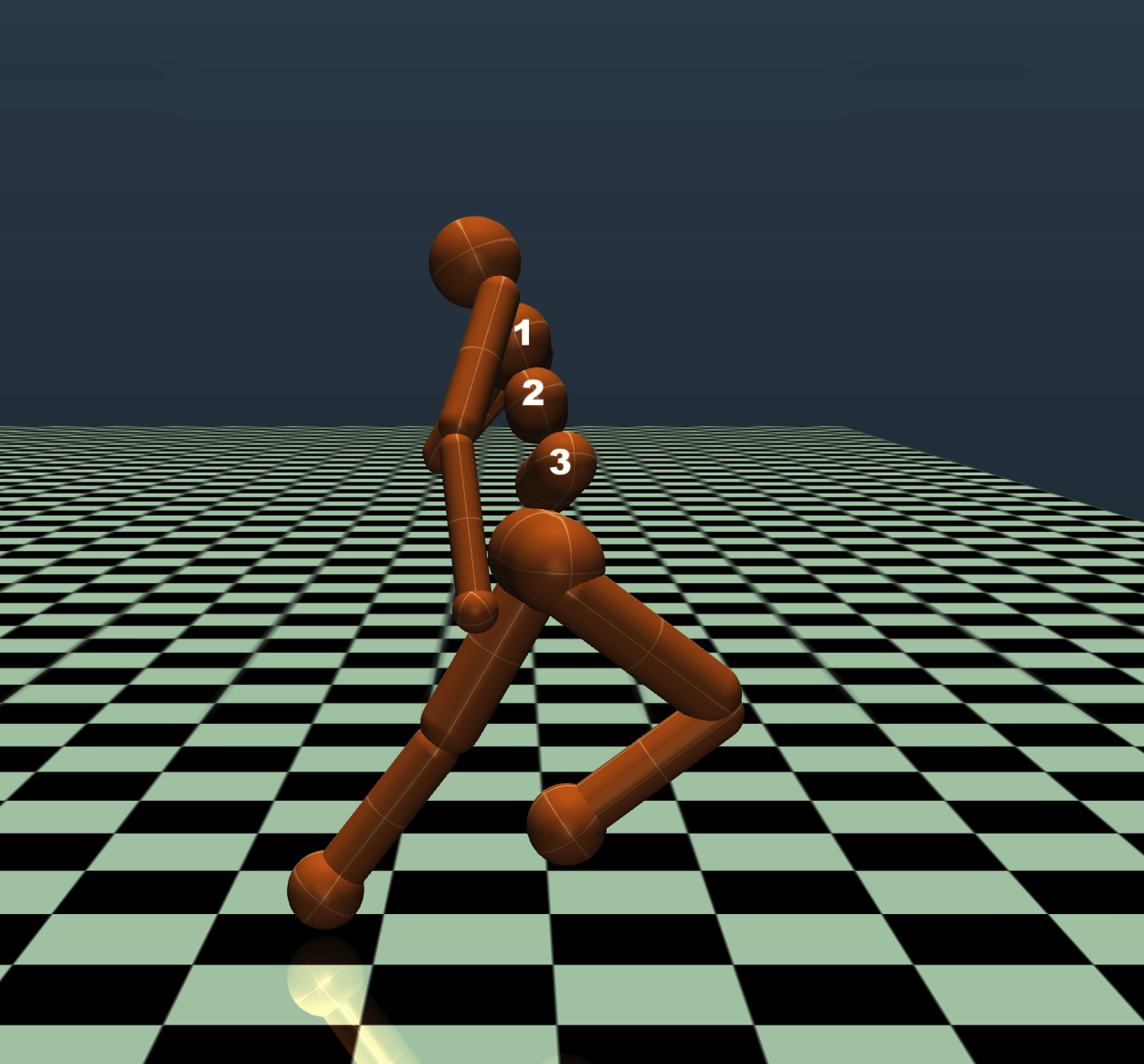}
	}
	\subfigure[]{
		\includegraphics[width=.600\textwidth]{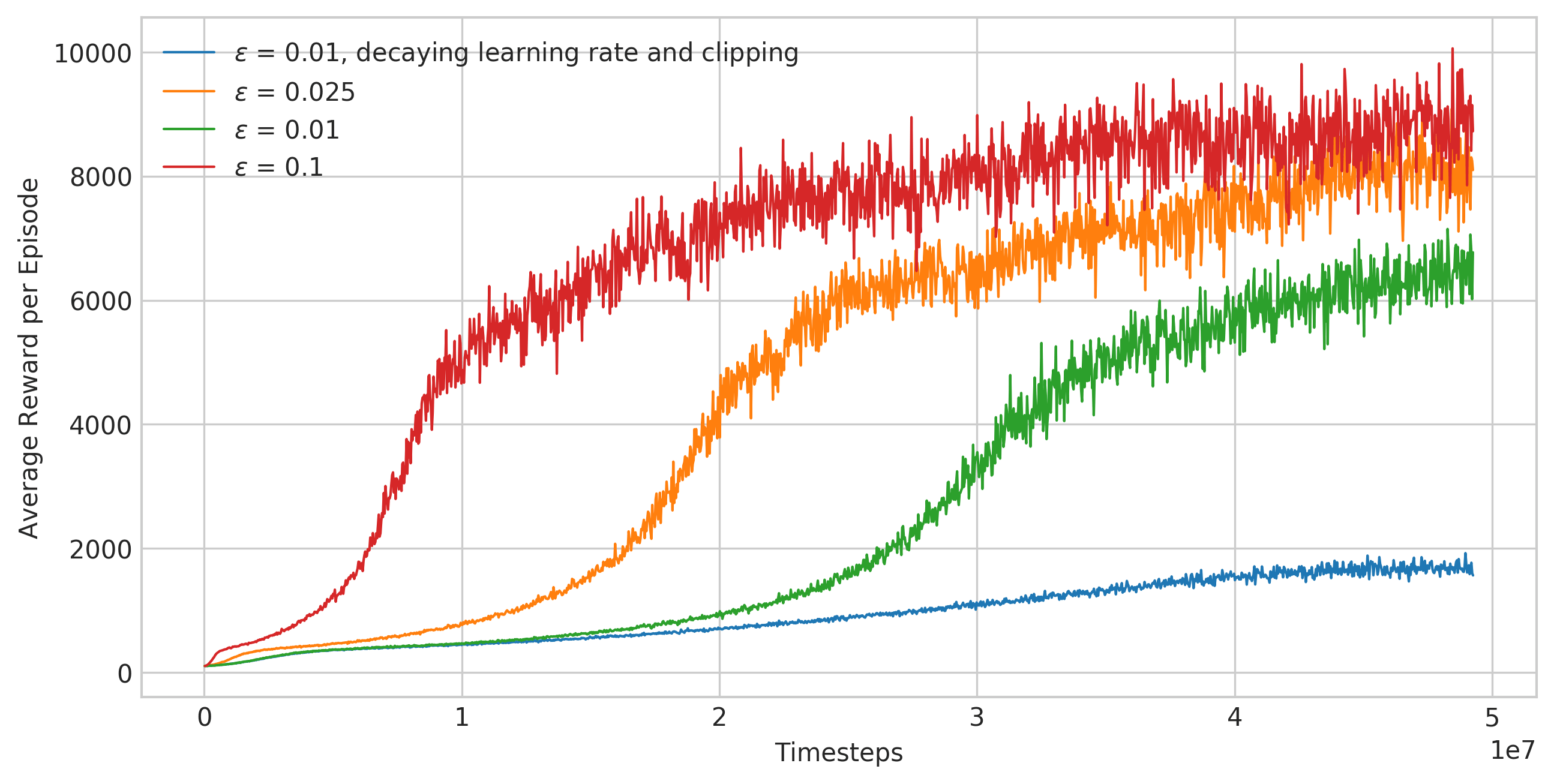}
	}
	\caption{(a) Humanoid running in the source environment. (b) Learning curves of SC-PPO and PPO on standard humanoid source task.}
	\label{fig:humsource}
\end{figure}

The average reward per episode and standard deviations of the policies trained with 4 different sets of hyperparameters are shown in Figure \ref{fig:humsource}(b). Learning curve obtained with the latest PPO hyperparameters suggested in \textit{OpenAI Baselines} framework \cite{baselines} for the Humanoid environment are represented by the red curve in Figure \ref{fig:humsource}(b). The learning curves for using strict clipping in the source tasks are shown with clipping hyperparameters 0.01, 0.025, and 0.01 decaying learning rate and clipping. 

%% file: files/sourcehopper.tex
\begin{figure}[!htbp]
	\centering
	\subfigure[]{
		\includegraphics[width=.330\textwidth]{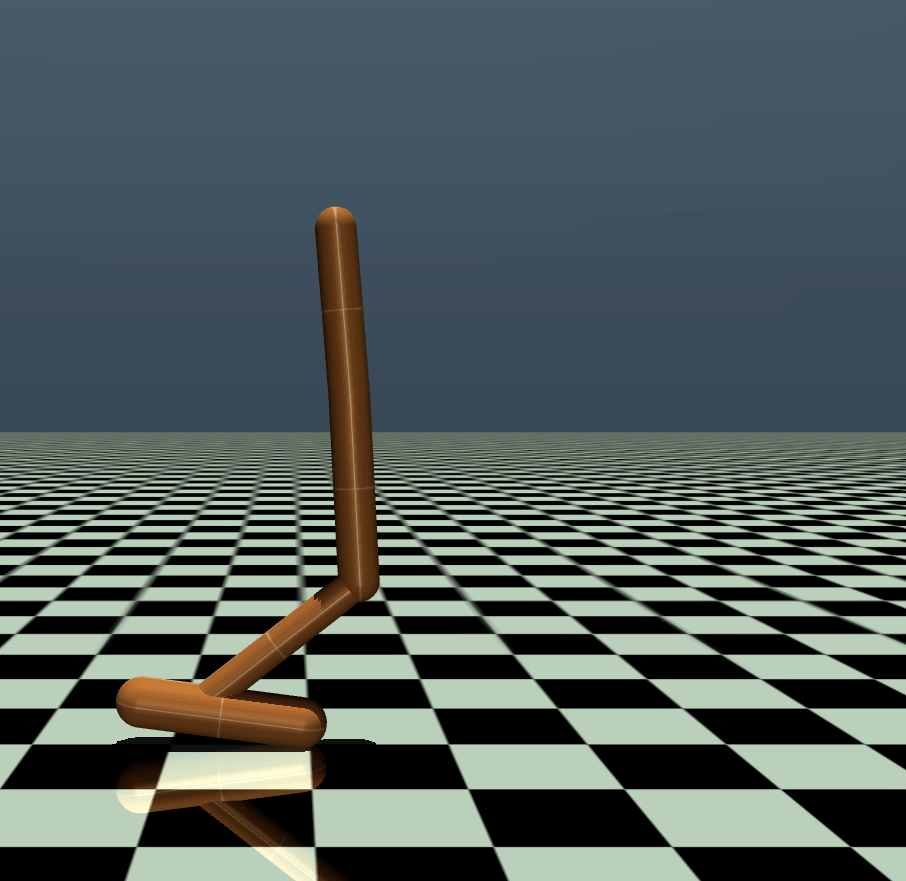}
	}
	\subfigure[]{
		\includegraphics[width=.500\textwidth]{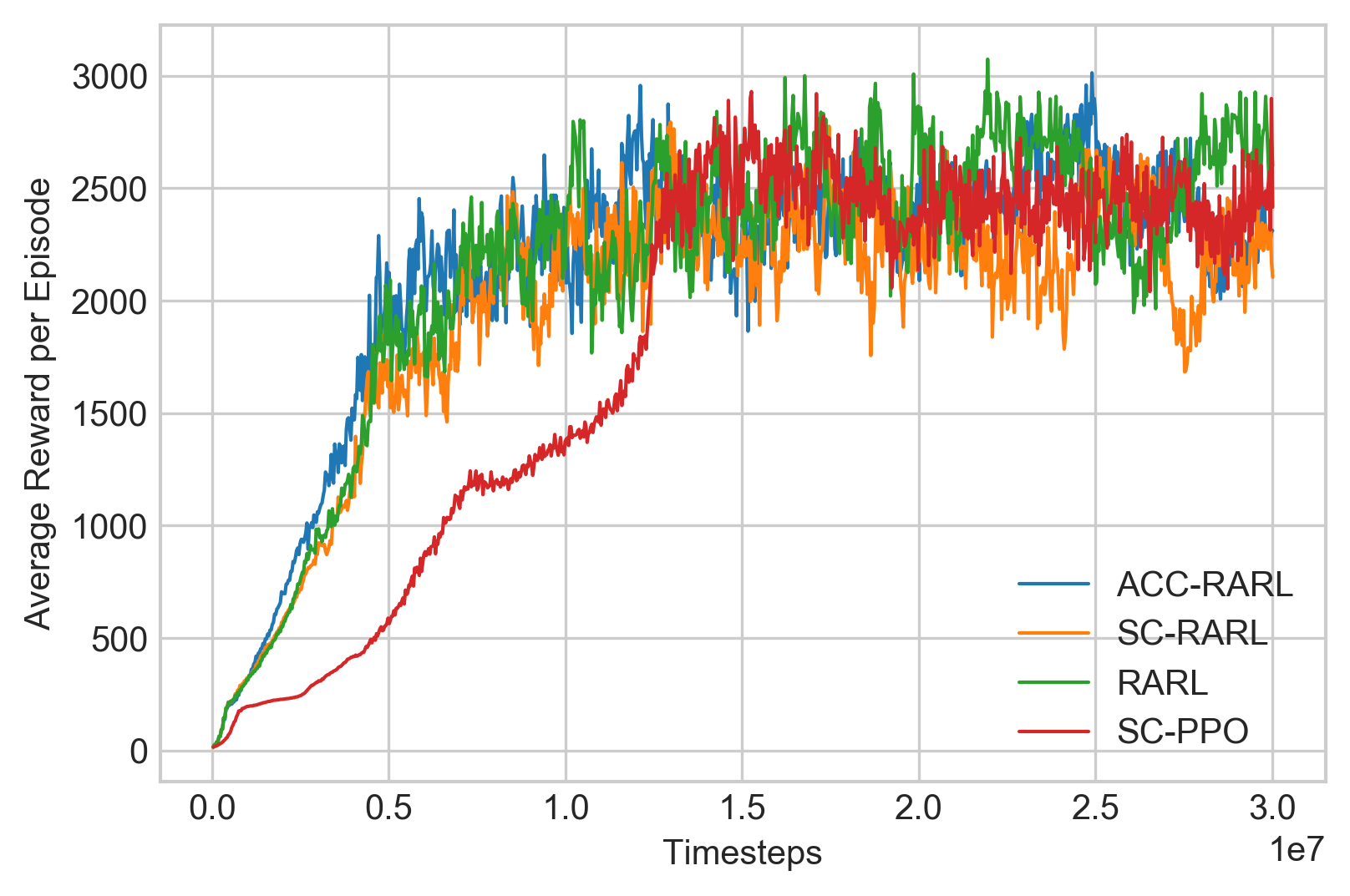}
	}
	\caption{(a) Hopping action, and (b) learning curves of ACC-RARL, SC-RARL, RARL, SC-PPO on standard hopper source task.}
	\label{fig:hoppersource}
\end{figure}
The Reward function of the hopper task consists of linear forward velocity reward, sum of squared actions and an alive bonus $C_{bonus}$.
\begin{equation}
\begin{aligned}
r _ { hopper } ( s , a )= \begin{array} { l l } {r _ {v _ { f w d }} - 0.001* \sum{ a^{2}} + C _ {bonus} }\end{array} 
\end{aligned}
\label{eq:HopperRwdFunction}
\end{equation}
SC-PPO was trained using $\alpha=0.0003$, $\epsilon=0.05$, $b=2048$, and the adversarial algorithms were trained with $\alpha=0.0003$, $\epsilon=0.3$, $b=512$. Source task learning curves of PPO and RARL trained with different critic architectures are shown in Figure \ref{fig:hoppersource}(b). Results in hopper tasks are consistent with the humanoid experiments. To analyze the effect of different architectures on the generalization capacity we compare three different critic architectures: separate double critic networks used in RARL \cite{rllab-adv}, single critic network in Shared Critic Robust Adversarial Reinforcement Learning (SC-RARL) \cite{rarlbaselines} and ACC-RARL. The policies trained with different variations of RARL perform similar to SC-PPO in hopper morphology tasks in the range of 1-6 and hopper gravity tasks in the range of 1G-1.5G. Next, we compare performances of these policies in various target tasks.

%% file: files/shorttall.tex
\begin{figure}[!htbp]
	\centering
	\subfigure[]{
		\includegraphics[width=.196\textwidth]{stdhum}
	}
	\subfigure[]{
		\includegraphics[width=.220\textwidth]{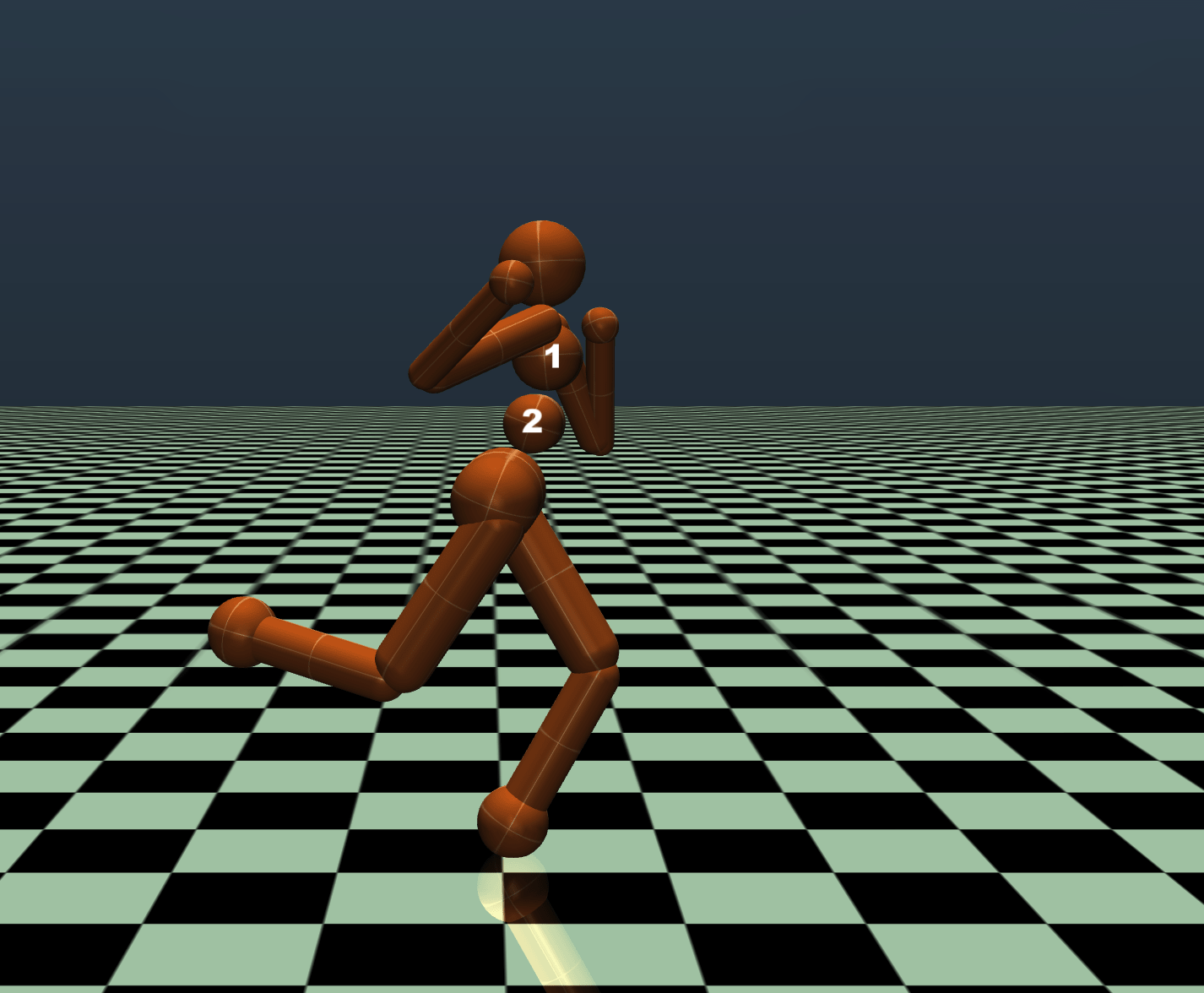}
	}
	\subfigure[]{
		\includegraphics[width=.232\textwidth]{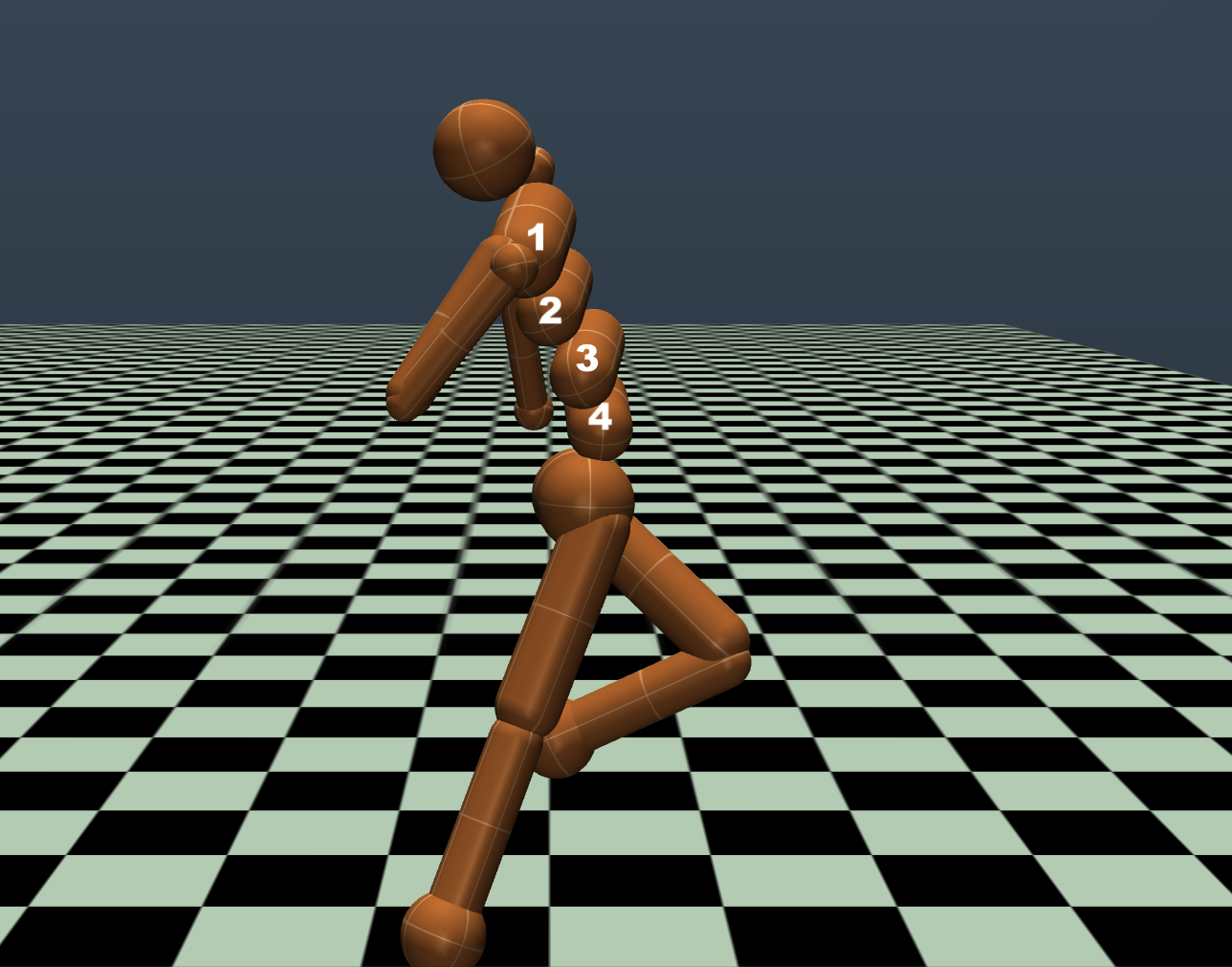}
	}
	\subfigure[]{
		\includegraphics[width=.220\textwidth]{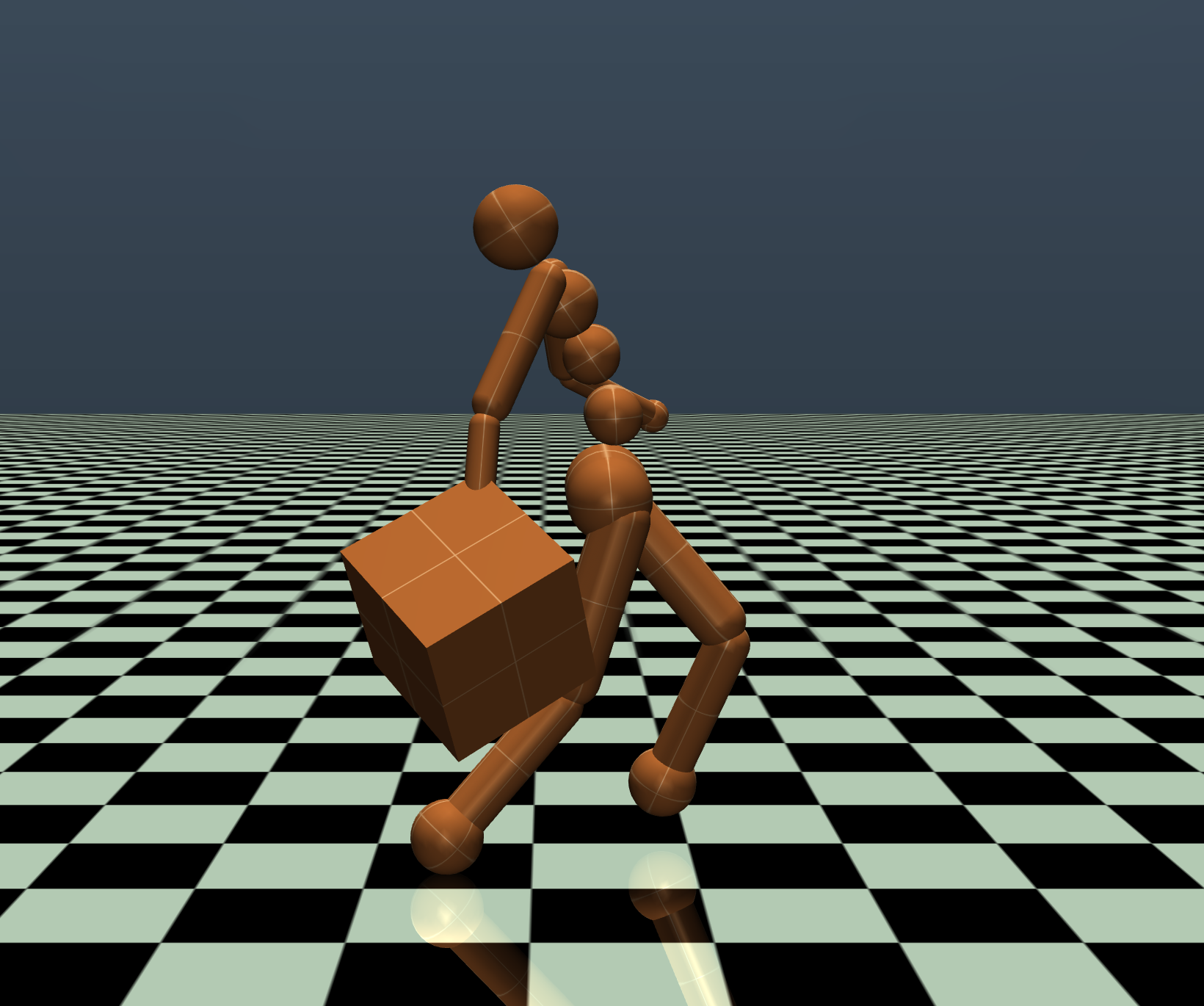}
	}
	\caption{(a) Standard humanoid source task with 3 torso components, (b) short humanoid target task with 2 torso components, (c) tall humanoid target task with 4 torso components, and (d) delivery humanoid target task}
	\label{fig:alltarghumpicture}
\end{figure}

The morphological target tasks are created for inter-robot transfer learning. The short, tall, and delivery humanoid target tasks in Figures \ref{fig:alltarghumpicture}(b), (c), (d) are generated from the standard humanoid source task in Figure \ref{fig:alltarghumpicture}(a).

The total body weights of the short and tall humanoid target tasks differ from the humanoid source task by the exclusion and inclusion of the upper waist respectively. The tall humanoid task is more challenging than the short humanoid task because it is harder to balance heavier upper body mass where the center of mass is higher from the ground. The termination criterion for bipedal locomotion in humanoids depends on the location of the center of the torso. When the +z location of the center of mass of the robot is below a threshold, the robot is presumed to fall and the episode is terminated. This threshold is higher for the tall humanoid and lower for the short humanoid because of the different locations of the center of mass.

\counterwithin{table}{section}
\label{table:delivenvspecs}
\begin{table}[!htbp]
	\vskip\baselineskip 
	\caption[Delivey Mass]{Delivery Humanoid Environment}
	\begin{center}
		\begin{tabular}{|c|c|}\hline
			\textbf{Body}& \textbf{Unit Mass}\\\hline
			Delivery Box & 5\\\hline    
			Right Hand  & 1.2\\\hline
			Torso & 8.3\\\hline
			Total Body without Delivery Box& 40.0\\\hline
		\end{tabular}
	\end{center}
\end{table}

Masses of the relevant body parts for the delivery humanoid are given in \ref{table:delivenvspecs}. Considering the total body mass of 40, a box with a unit mass of 5 constitutes a challenging benchmark. Our purpose is to create a horizontal imbalance by enforcing the humanoid to carry the box only by the right hand. 

\begin{figure}[!htbp]
	\centering
		\subfigure[]{
		\includegraphics[width=.480\textwidth]{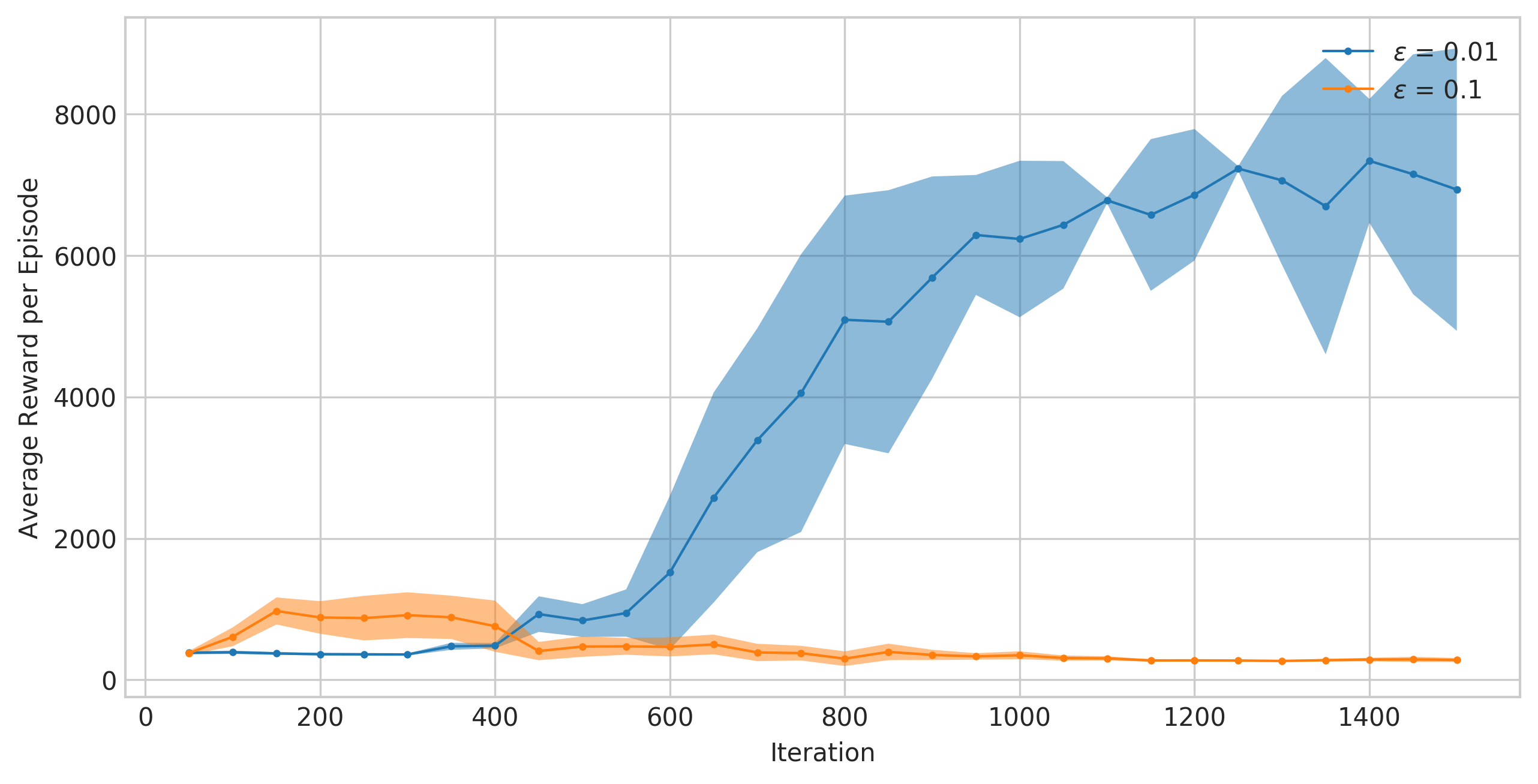}
		\label{fig:humshort}
	}
	
	\subfigure[]{
		\includegraphics[width=.480\textwidth]{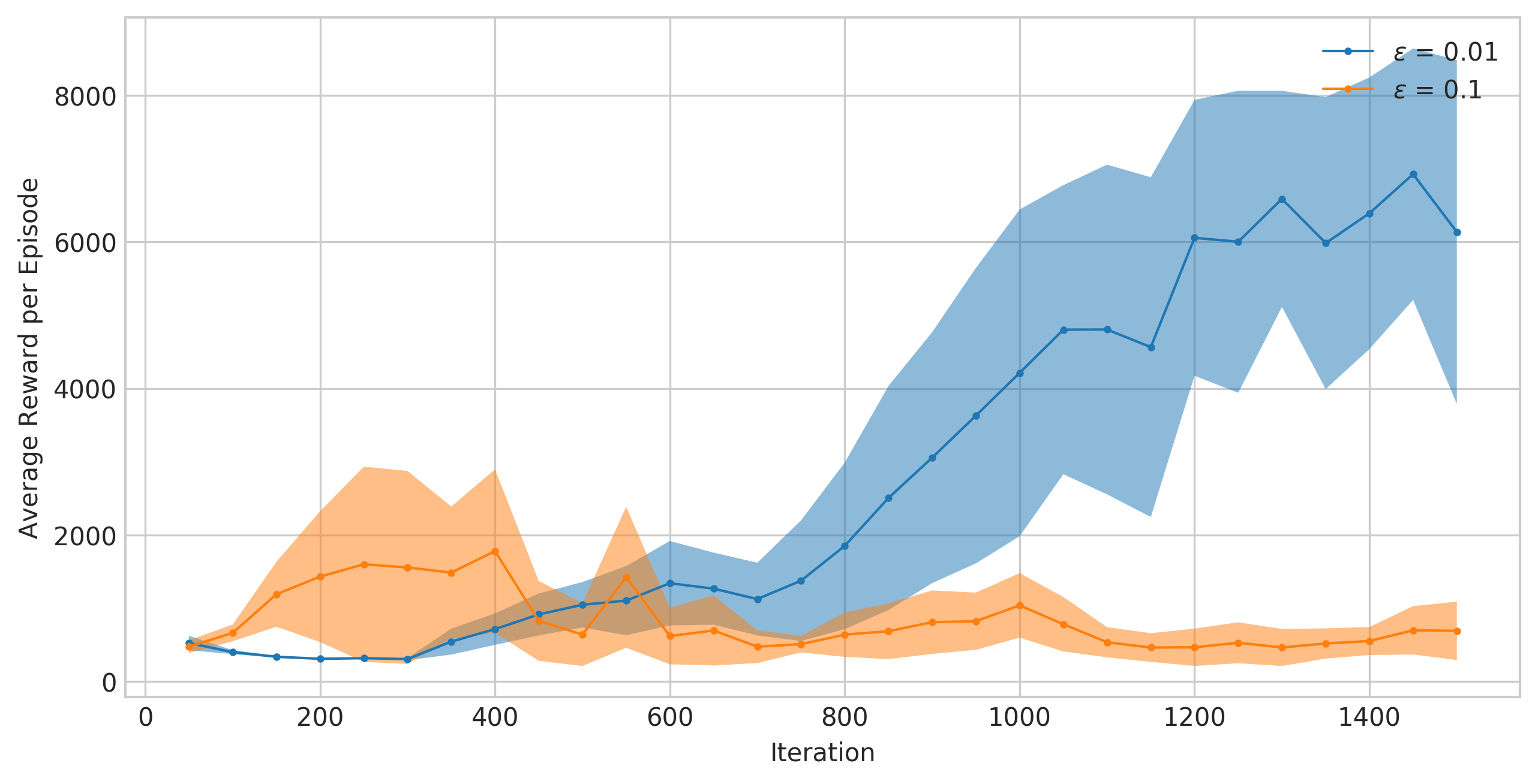}
			\label{fig:humtall}
	}
		\subfigure[]{
		\includegraphics[width=.480\textwidth]{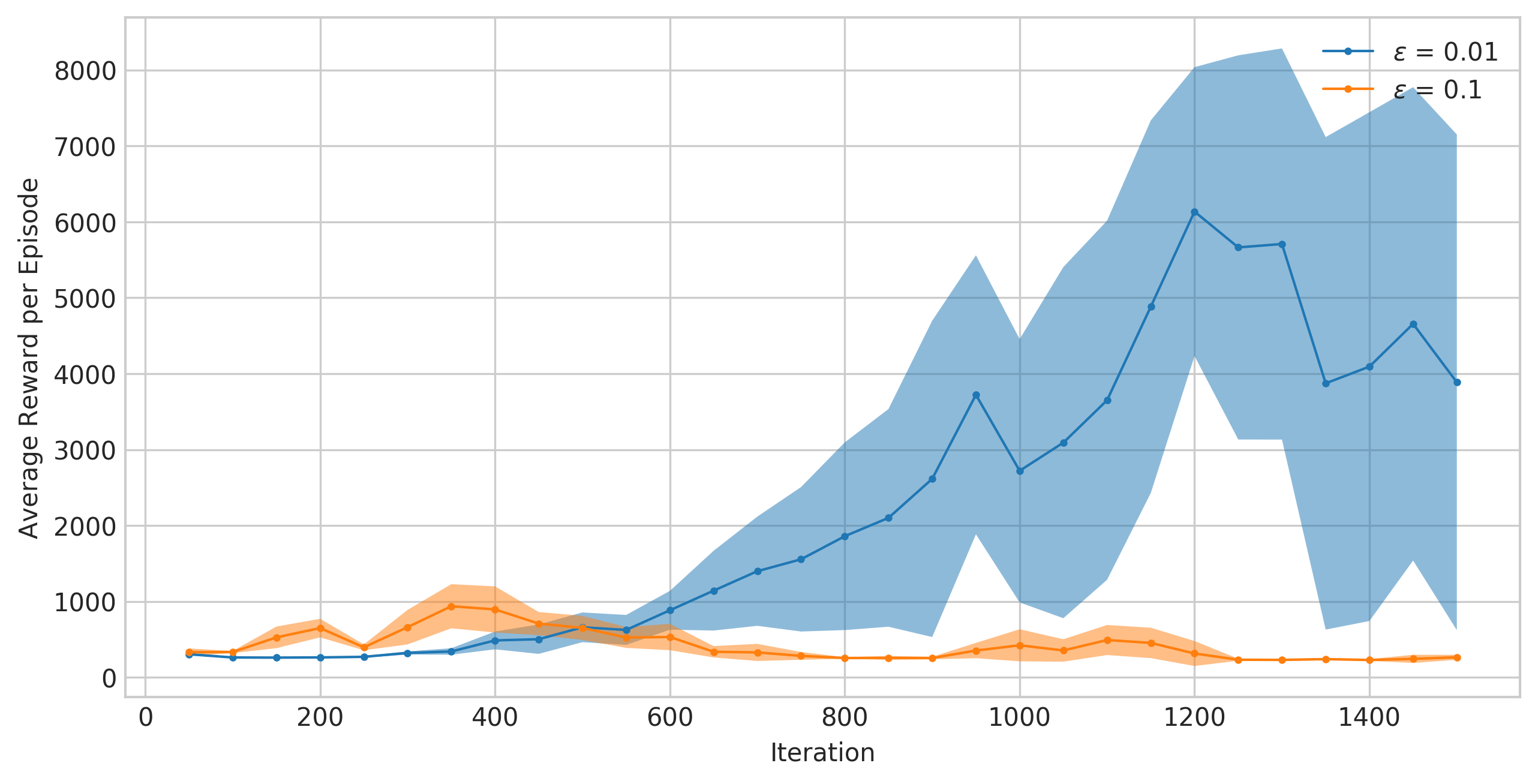}
		\label{fig:humhandbox}
	}
	\caption{Performance of SC-PPO and PPO on (a) shorter and lighter humanoid target task, (b) taller and heavier humanoid target task, and (c) delivery humanoid target task.}
\end{figure}

The comparison of the target task performances of policies trained with clipping parameters $\epsilon=0.1$ and $\epsilon=0.01$ are provided in Figures \ref{fig:humshort}, \ref{fig:humtall} and \ref{fig:humhandbox}. High average reward per episode obtained after $1200^{th}$ iteration with SC-PPO $\epsilon=0.01$ show that humanoid learns transferable general characteristics of forward locomotion from the source task. In contrast, all iterations of the policies trained with clipping parameter $\epsilon=0.1$ failed in all target tasks. Early stopping regularization technique alone is not sufficient because the short and tall humanoids can not stand still using the earlier iterations of the policy trained with clipping parameter $\epsilon=0.1$.

The reduction in the performance after the $1200^{th}$ iteration in Figure \ref{fig:humhandbox}, supports the method of resorting to the earlier policy iterations. This concavity suggests that overfitting occurs and early stopping is an effective regularization technique.

The same policy iteration ($1200^{th}$) trained with SC-PPO can be successfully transferred to short, tall and delivery humanoid tasks as shown in Figures \ref{fig:humshort}, \ref{fig:humtall} and \ref{fig:humhandbox}. Evaluation of the best source task policies in the target tasks assesses our claim that source task performance is not indicative of the generalization capacity. Thus, all transfer RL methods should account for regularization first.

%% file: files/fric_exp.tex
Our aim in this experiment is to evaluate our methods in transfer learning benchmarks where the environment dynamics are changed. We generate a target task by increasing the ground friction coefficient in MuJoCo environment. The humanoid sinks due to high tangential friction but can still run following the regularized policies trained with SC-PPO. 

\begin{figure}[!htbp]
	\centering
	\subfigure[]{
		\includegraphics[width=.570\textwidth]{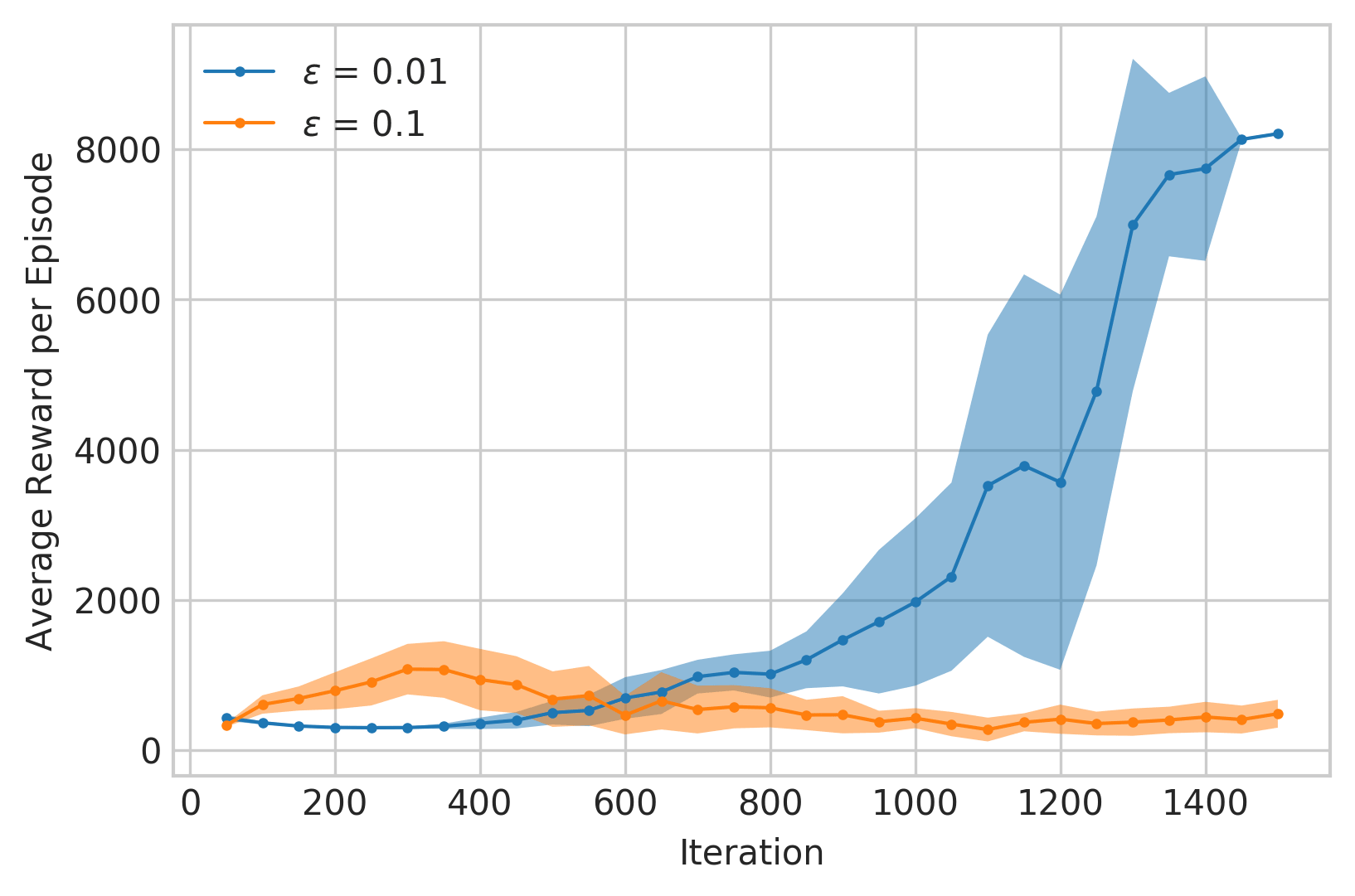}
	}
	\caption{Performance of SC-PPO and PPO on target environment with tangential friction 3.5 times the source environment.}
	\label{fig:fric}
\end{figure}

\begin{table}[!htbp]
	\caption[Friction Environment]{Comparison of SC-PPO and PPO in Target Friction Environment}
	\begin{center}
		\begin{tabular}{|c|c|c|}\hline
			\textbf{Clip}& \textbf{Iteration}&\textbf{Average Reward per Episode}\\\hline
			0.01  & 1500 &$8283 \pm 24.3$\\\hline
			0.1 &300 &$1078 \pm 336.4$\\\hline    
		\end{tabular}
		\label{table:fricenvspecs}
	\end{center}
\end{table} 

The best jumpstart performances for each clipping parameter are given in Figure \ref{fig:fric}. For instance, the last iteration of the policy with a SC-PPO parameter $\epsilon=0.01$ has an average target task reward of 8283 with 24.3 standard deviation as provided in Table \ref{table:fricenvspecs}. In contrast, the best performing policy in the source task has a target task performance which assesses our claim that source task performance on transfer RL is not indicative of the target task performance. These results show that agent trained using our methods learns generalizable skills for environments with changing dynamics. 

%% file: files/grav_hum.tex
Our aim in these experiments is to evaluate our methods in a different set of target tasks where environment parameters are changed. Similar to the friction experiments, we generate gravity target tasks in the range of 0.5G-1.75G where G=-9.81 is the gravity of the source task.

\begin{figure}[!htbp]
	\centering
	\subfigure[]{
		\includegraphics[width=.600\textwidth]{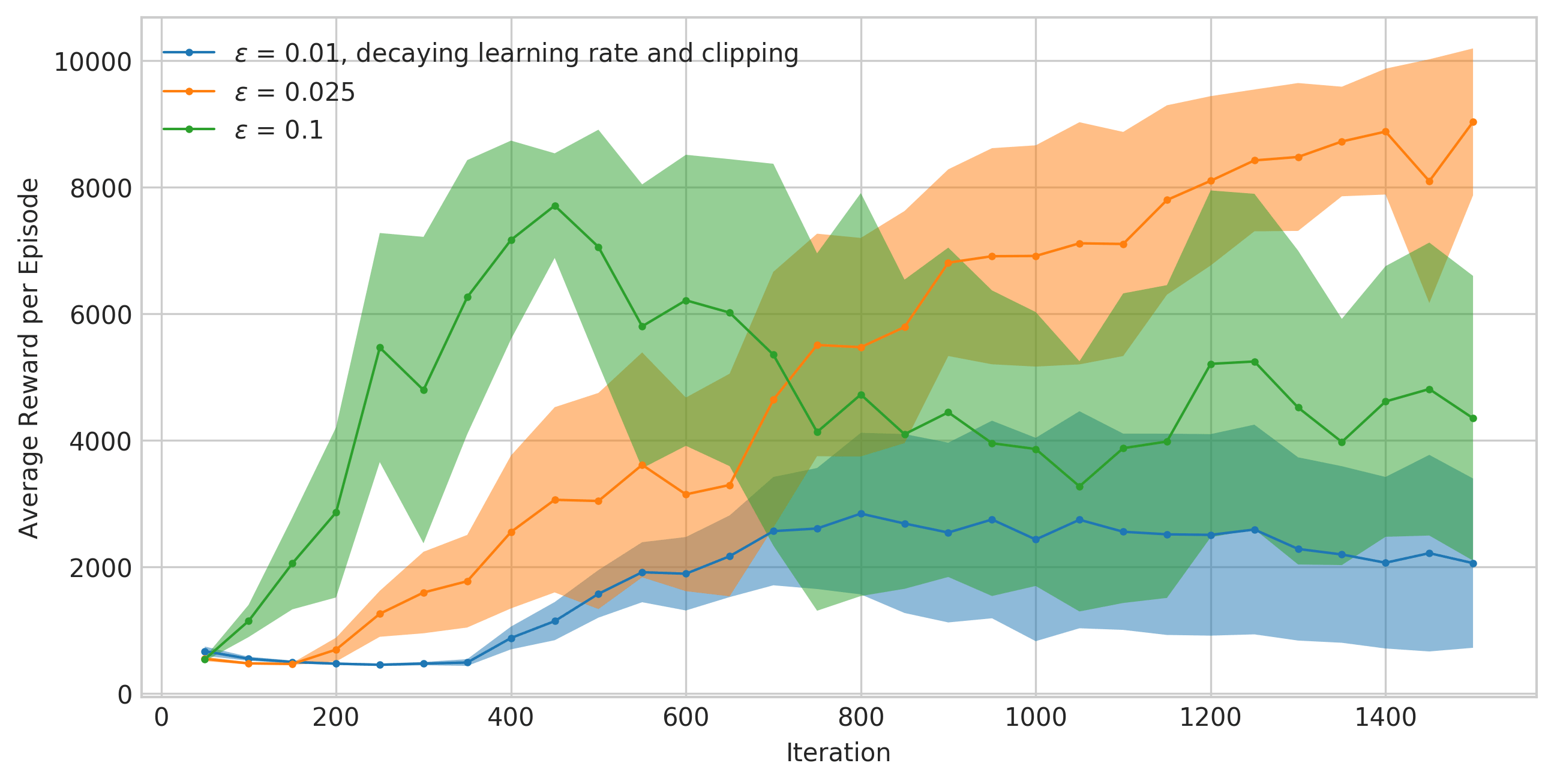}
	}
	\caption{Performance of SC-PPO and PPO on target environment with gravity= -4.905 ($0.5G_{Earth}$). }
	\label{fig:grav4}
\end{figure}

\begin{figure}[!htbp]
	\centering
	\subfigure[]{
		\includegraphics[width=.600\textwidth]{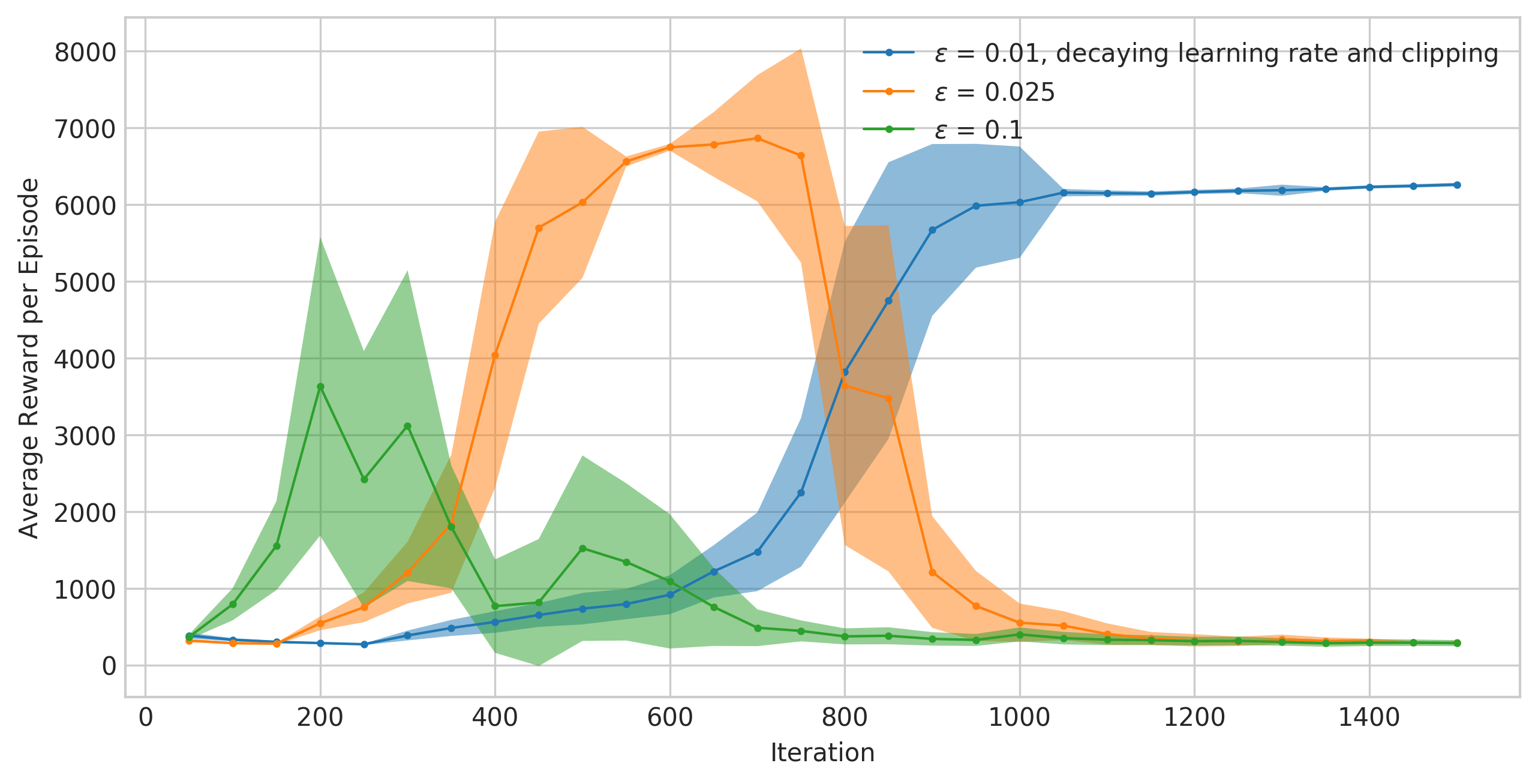}
	}
	\caption{Performance of SC-PPO and PPO on target environment with gravity= -14.715 ($0.5G_{Earth}$).}
	\label{fig:grav14}
\end{figure}

When the last iteration of the policy trained with SC-PPO $\epsilon=0.025$ is used in the target task with gravity= -4.905 ($0.5G_{Earth}$) the humanoid in Figure \ref{fig:grav4}(a) can run. Performance of SC-PPO is slightly better than early stopping for this target task. Similarly, in the target task with gravity= -14.715 ($1.5G_{Earth}$), the humanoid needs to resort to the previous snapshots of the policy trained with SC-PPO $\epsilon=0.025$ as plotted in the Figure \ref{fig:grav14}(a). The bipedal locomotion pattern in the simulated target environment with gravity= -14.715 ($1.5G_{Earth}$) when the humanoid jumpstarts with the $600^\textsuperscript{th}$ policy trained with clipping parameter $\epsilon=0.025$ is shown in Figure \ref{fig:grav14}(b).

\begin{figure}[!htbp]
	\centering
	\subfigure[]{
		\includegraphics[width=.600\textwidth]{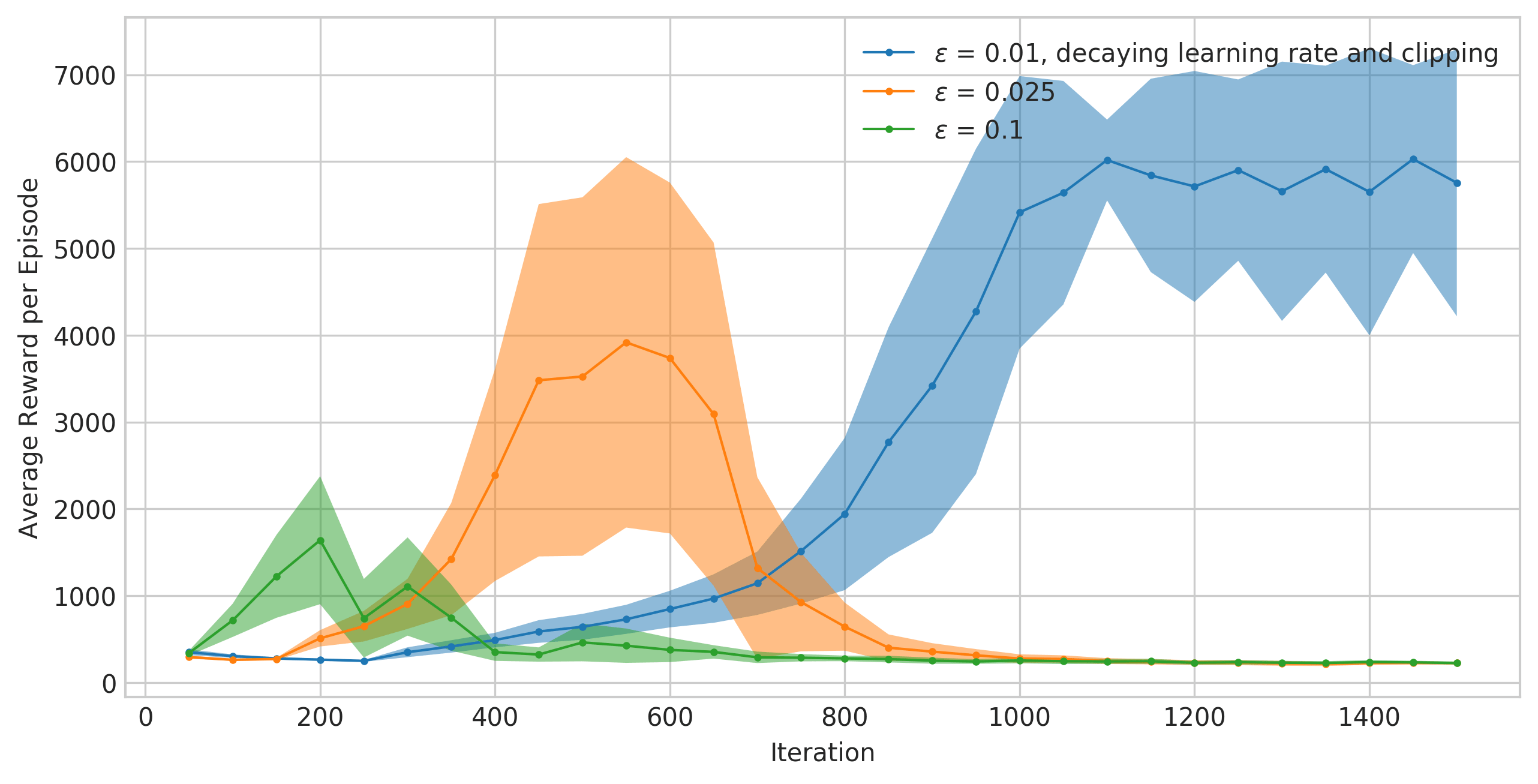}
	}
	\caption{Performance of SC-PPO and PPO on target environment with gravity= -17.1675 ($1.75G_{Earth}$).}
	\label{fig:grav17}
\end{figure}

Gravity benchmarks for the humanoid indicate that snapshots of different policies should be used for the target task with gravity= - 17.1675 ($1.75G_{Earth}$). The policy iterations trained with hyperparameters ``\textit{$\epsilon=0.01$ decaying learning rate and clipping}" performed poorly in the source task and target task with lower gravity. However, they perform consistently well in environments with a higher magnitude of gravity (- 17.1675,- 14.715). Figures \ref{fig:grav14} and \ref{fig:grav17} show that decaying the clipping parameter and the learning rate during training decreased the exploration and restricted the humanoid to stick to a more careful way of moving forward after 1000 training iterations. 

%% file: files/torsomasshopper.tex
In this set of experiments, we compare our methods with RARL. In RARL, target tasks are generated by modifying the torso mass of the robot in the range of 2.5-4.75 \cite{pinto2017robust}. Following the same procedure, we modify the torso mass in the range of 1-9. Table \ref{table:hopperspecs} provides additional information on the morphology of the Hopper source task.

\begin{table}[!htbp]
	\vskip\baselineskip 
	\caption[Hopper Mass]{Hopper Source Environment}
	\begin{center}
		\begin{tabular}{|c|c|}\hline
			\textbf{Body}& \textbf{Unit Mass}\\\hline
			Torso & 3.5\\\hline    
			Thigh & 3.9\\\hline
			Leg & 2.7\\\hline
			Foot & 5.1\\\hline
			Total Body & 15.3\\\hline
		\end{tabular}
	\end{center}
	\label{table:hopperspecs}
\end{table} 

We evaluate the target task performances of RARL \cite{rllab-adv}, Shared Critic Robust Adversarial Reinforcement Learning (SC-RARL) \cite{rarlbaselines}, and ACC-RARL. The only difference between SC-RARL and RARL is that in SC-RARL, a shared critic network is used for value function approximation. No regularization is used in \cite{rarlbaselines}, thus its generalization capacity is similar to RARL. In our work, we first regularize all adversarial architectures, then compare their generalization capacity. 

\begin{figure}[!htbp]
	\centering
	\subfigure[]{
		\includegraphics[width=.480\textwidth]{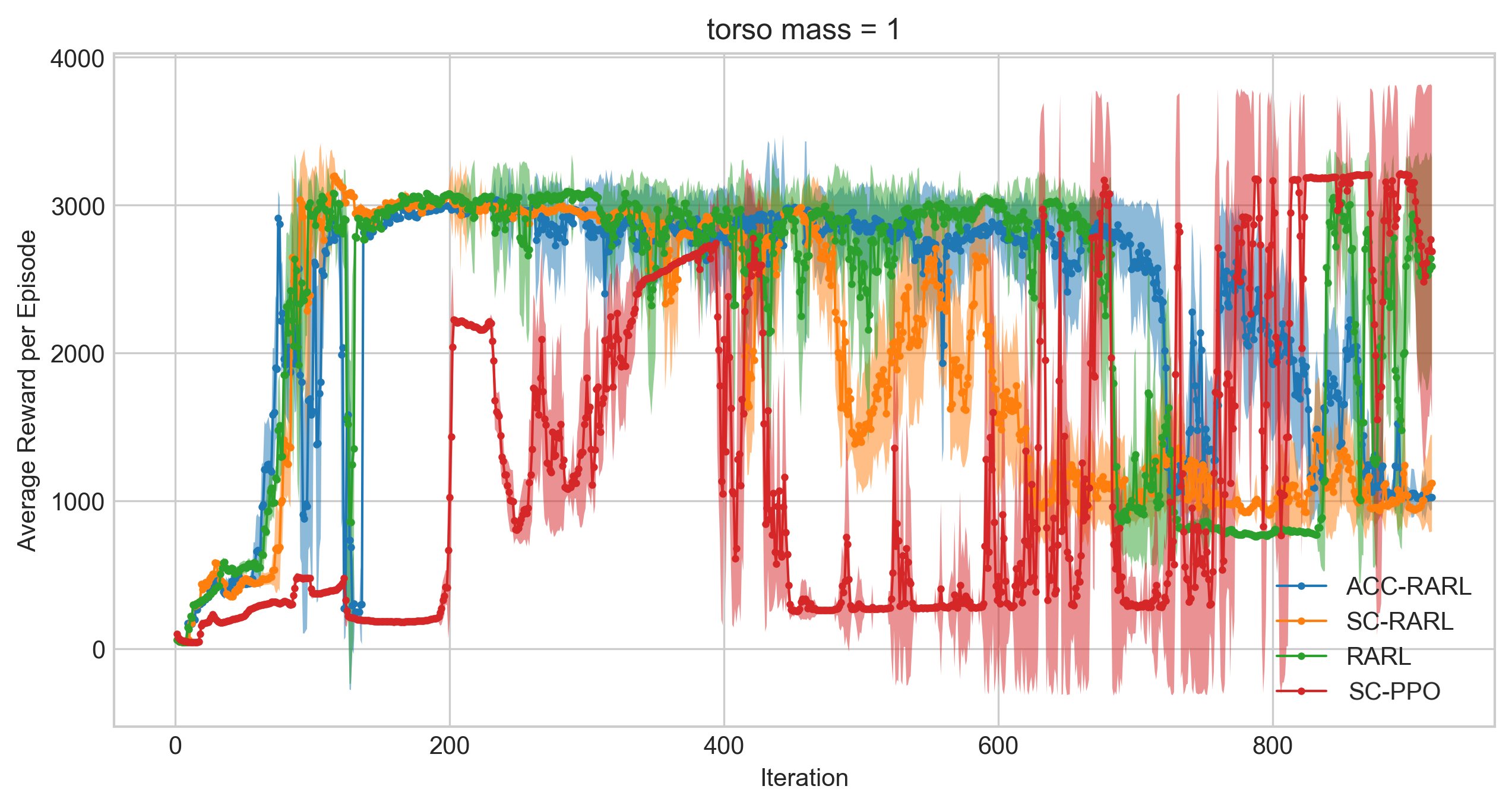}
	}
	\subfigure[]{
		\includegraphics[width=.480\textwidth]{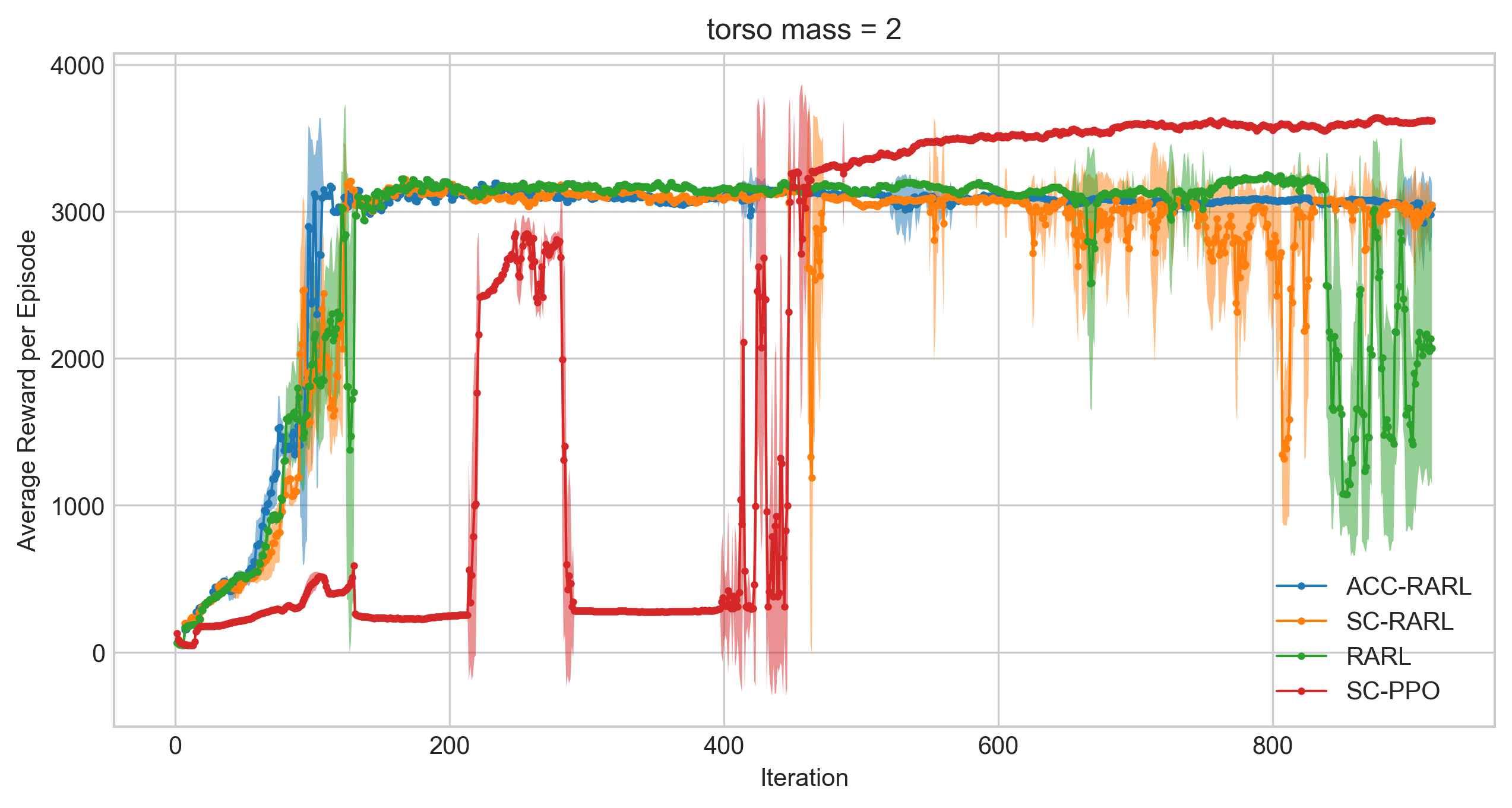}
	}
	\subfigure[]{
		\includegraphics[width=.480\textwidth]{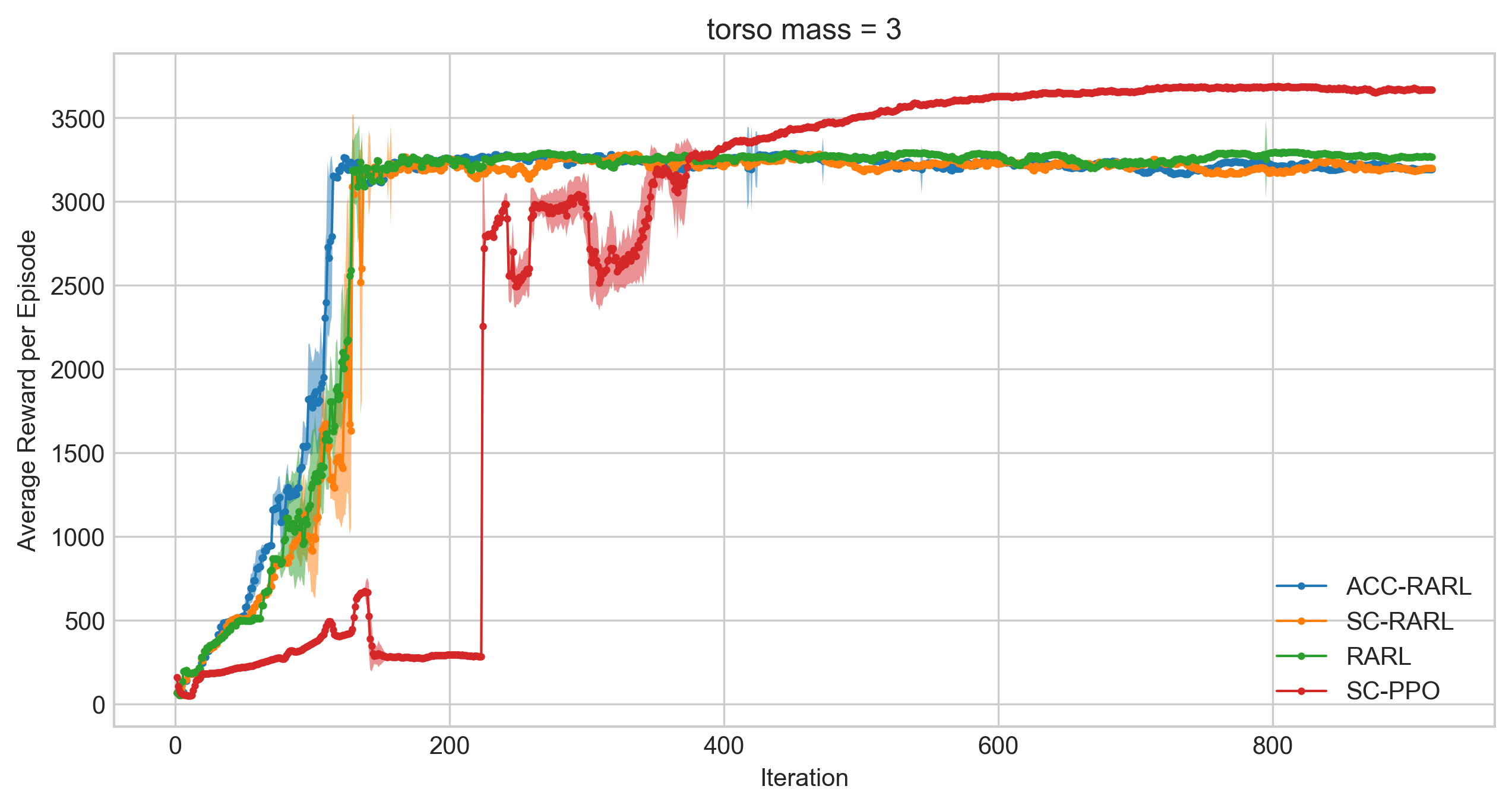}
	}
	\subfigure[]{
		\includegraphics[width=.480\textwidth]{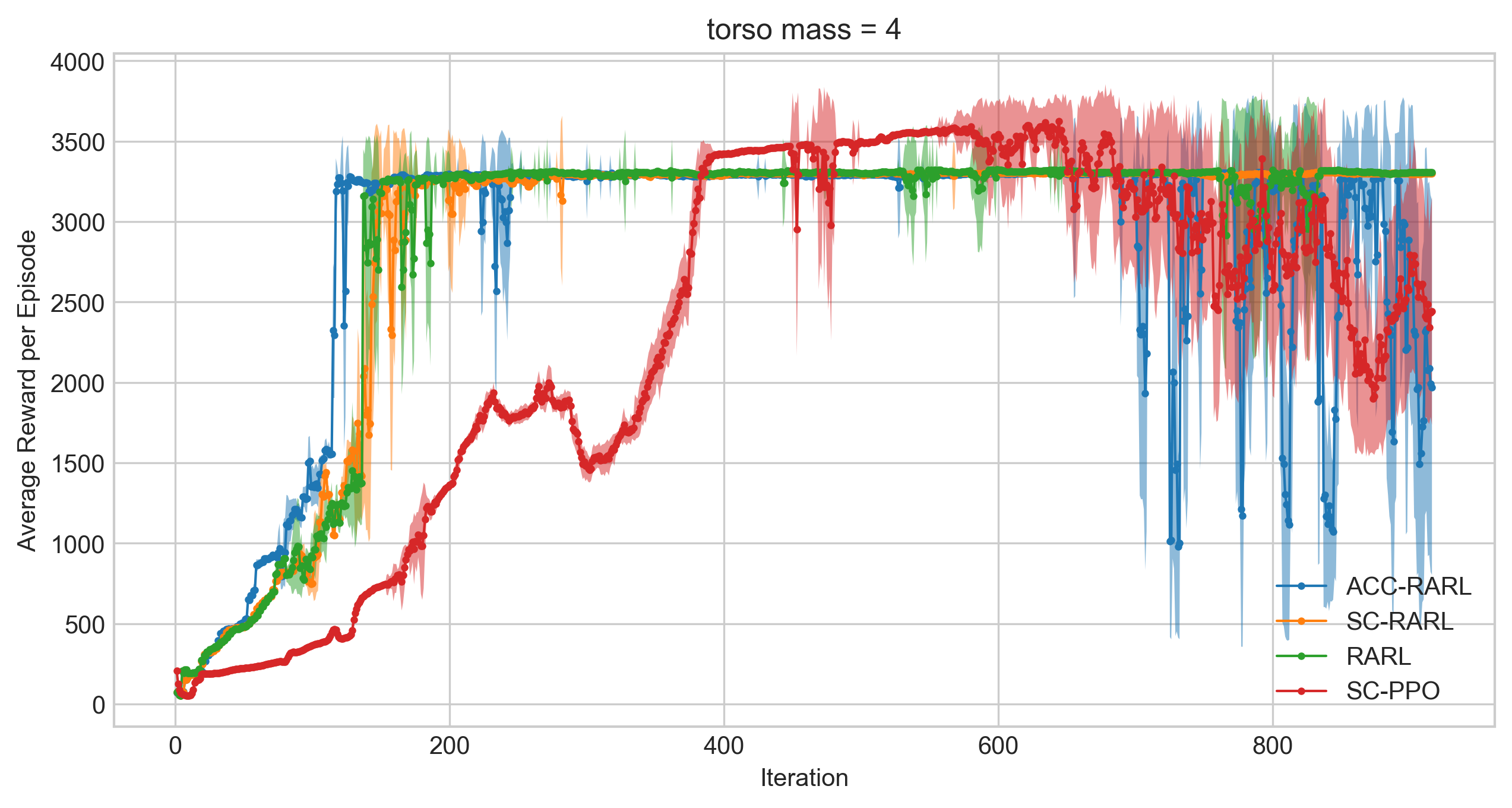}
	}
	\subfigure[]{
		\includegraphics[width=.480\textwidth]{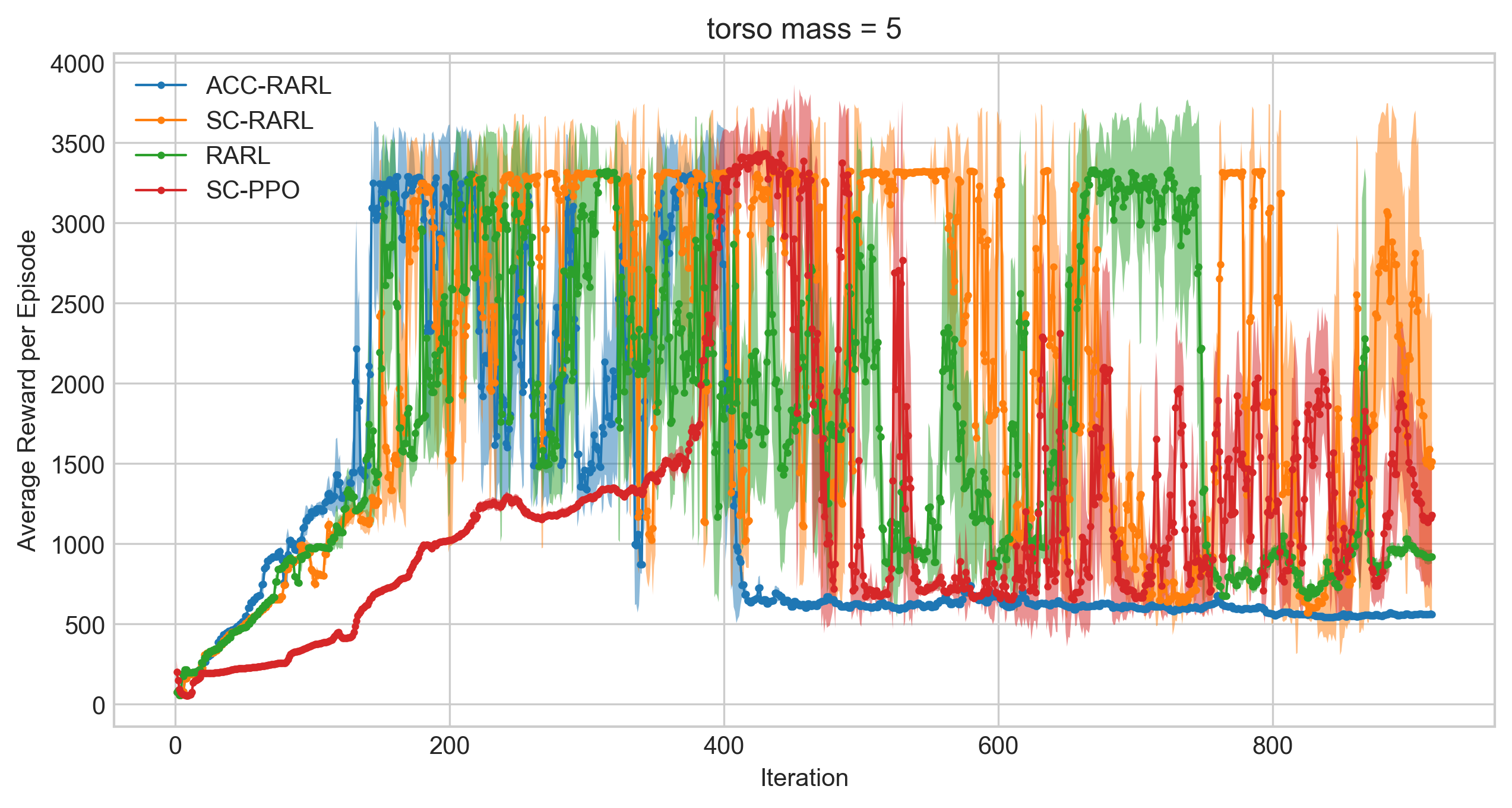}
	}
	\subfigure[]{
		\includegraphics[width=.480\textwidth]{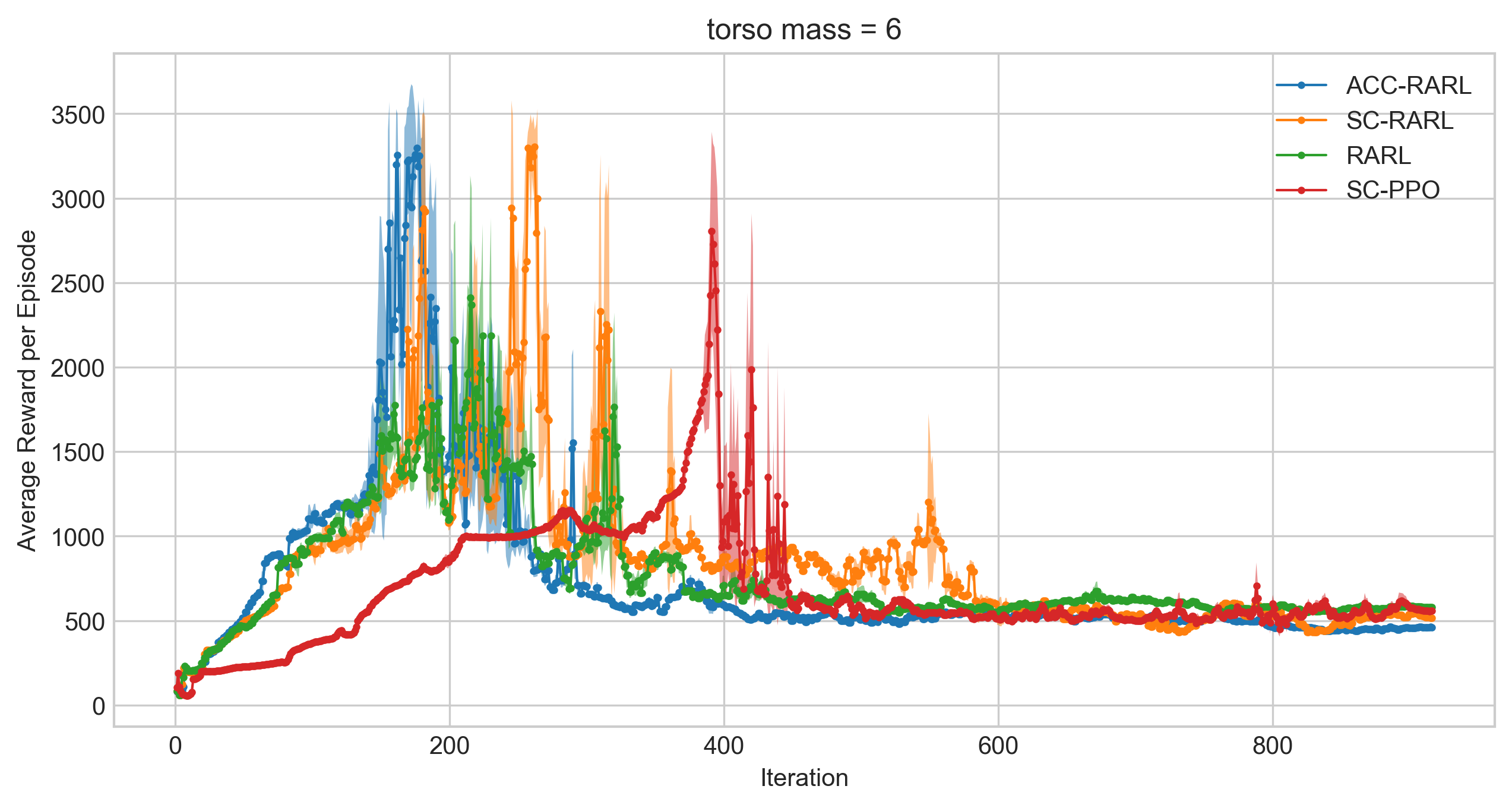}
	}
	\subfigure[]{
		\includegraphics[width=.480\textwidth]{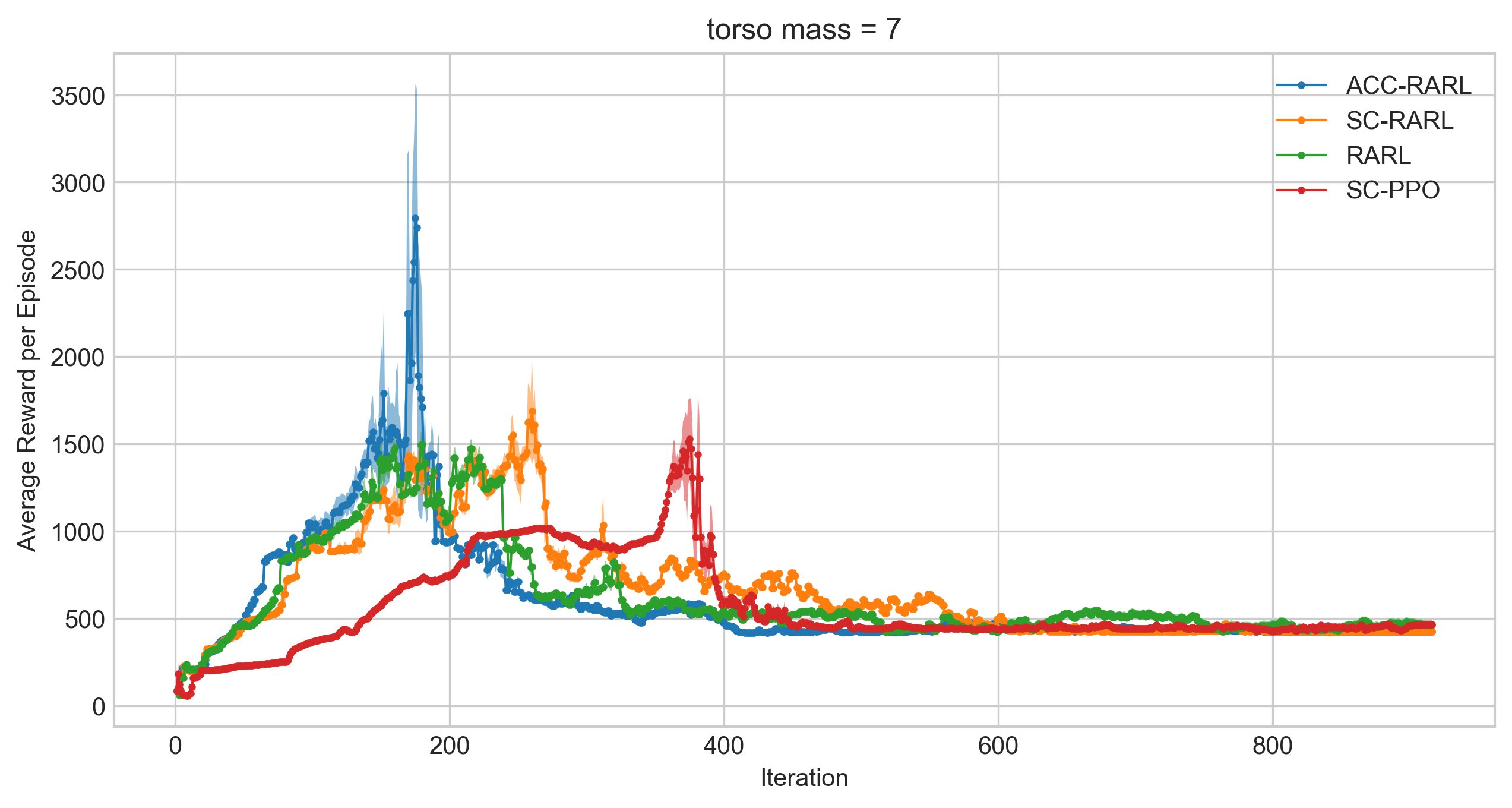}
	}
	\caption{Performance of SC-PPO and adversarial methods on Hopper target tasks with torso mass (a)1, (b)2, (c)3, (d)4, (e)5, (f)6, (g)7}
	\label{fig:hoppermassiter10}
\end{figure}

The robot can successfully generalize using any regularized adversarial technique or SC-PPO with early stopping when the target tasks are in the range of 1-6. Based on the target task performances of ACC-RARL, SC-RARL, and RARL, in Figure \ref{fig:hoppermassiter10}, different critic value function approximation techniques have an effect on the type of control behavior learned. However, the performance of some policy iterations trained with RARL, ACC-RARL, SC-RARL starts to become unstable in Figures \ref{fig:hoppermassiter10}(a) and \ref{fig:hoppermassiter10}(e). In harder target tasks like these, the agent should first resort to earlier snapshots of the policy and use early stopping for adversarial RL. Let us assume that the agent only has several snapshots of the policy in its buffer trained in the source task with standard torso mass. Then, the agent is put in a target task with torso mass=6 which is analogous to an agent expected to carry weight while performing a control task. We propose that in cases like these, instead of training from the very beginning, the agent should primarily resort to earlier policy iterations because the generalizable policies performing at the expert level are readily available in the agent's memory and were extracted from the source task training. 

If we had not recognized the policy iteration as a hyperparameter then comparing the algorithms at arbitrarily selected number of training iterations would not constitute a fair comparison. More importantly, the performance of PPO without adversaries, generally used as a benchmark algorithm, is highly dependent on the number training iterations. Hence for target tasks with torso masses $[1-6]$, the right snapshots of SC-PPO are capable of obtaining high average reward per episode as seen in Figures \ref{fig:hoppermassiter10}(a), \ref{fig:hoppermassiter10}(b), \ref{fig:hoppermassiter10}(c), \ref{fig:hoppermassiter10}(d), \ref{fig:hoppermassiter10}(e), \ref{fig:hoppermassiter10}(f). Furthermore, the skills learned with RARL in the last 150 iterations is successful in the source task and target task where torso mass is 3 and 4 units as provided in Figure \ref{fig:hoppermassiter10}(c). However, they are clearly unstable in tasks where torso mass is 1,2,5,6,7 as illustrated in Figures \ref{fig:hoppermassiter10}(a). The policy buffer allows us to do a fair comparison among different methods. Training the algorithms for an arbitrary number of iterations will produce inaccurate results such as underperforming baselines or proposed algorithms. Furthermore, the reproducibility of the transfer RL algorithms will be affected.

\begin{figure}[!htbp]
	\centering
	\subfigure[]{
	\includegraphics[width=.225\textwidth]{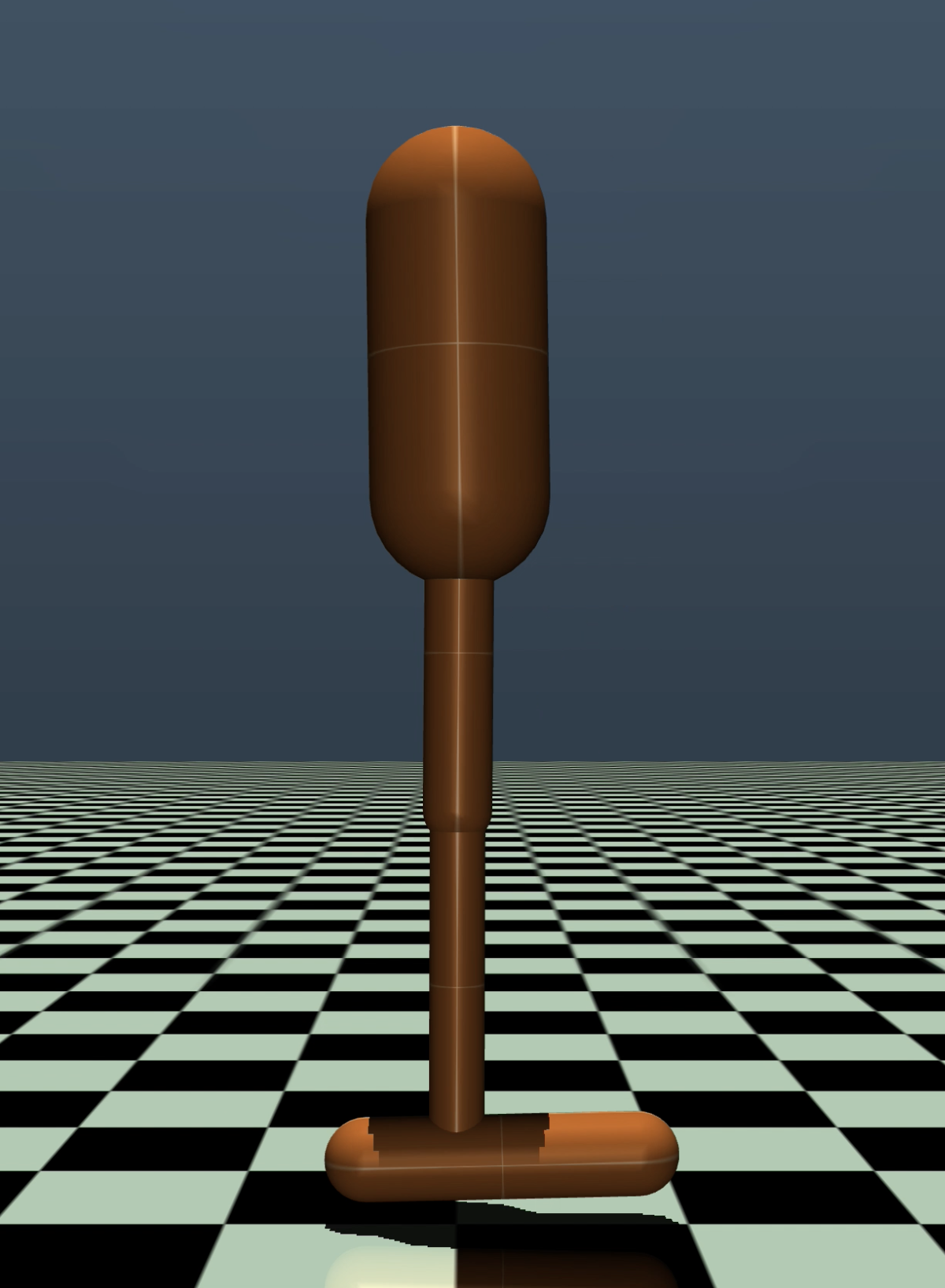}}
	\subfigure[]{
	\includegraphics[width=.600\textwidth]{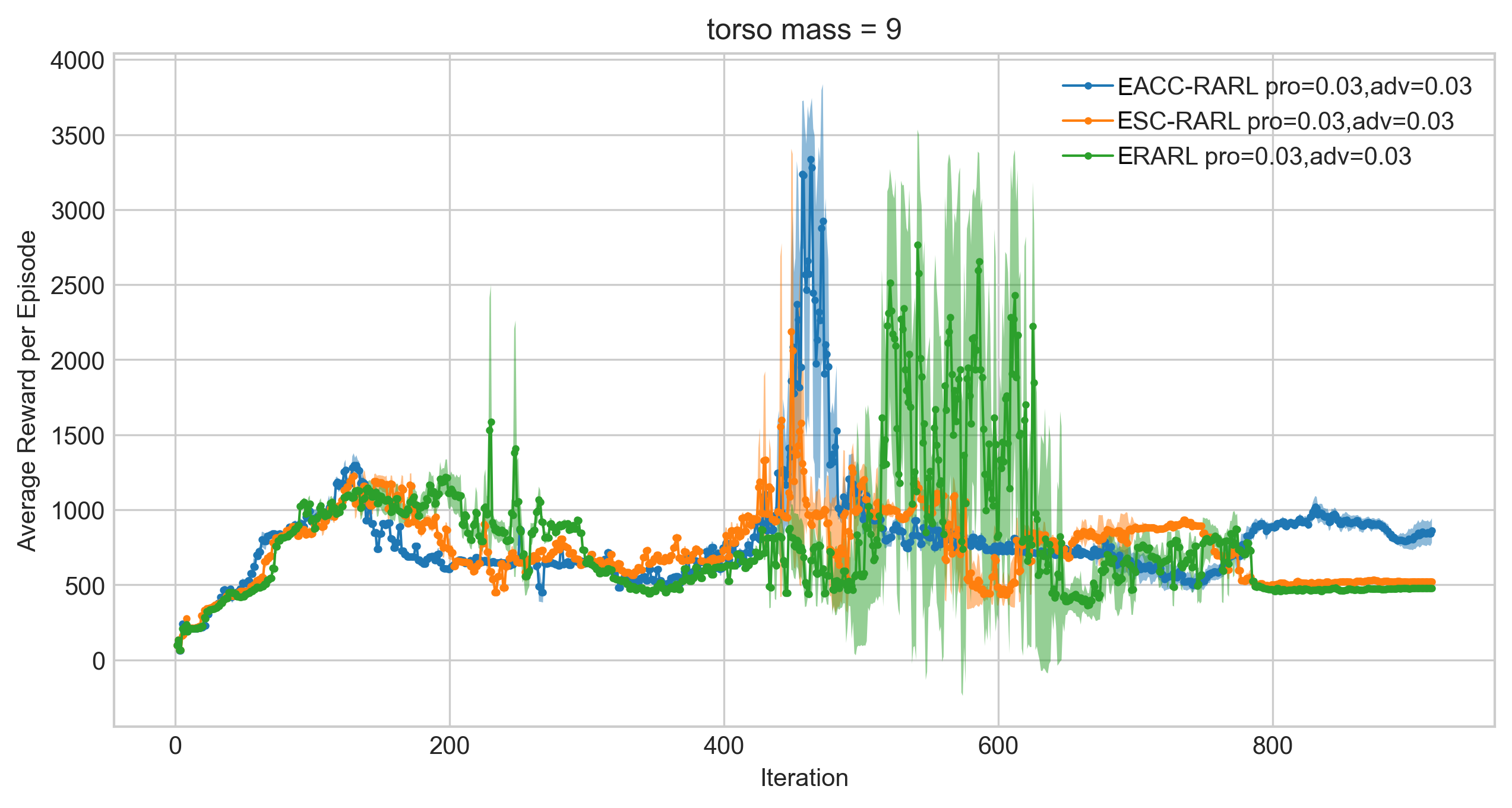}}
	\caption{(a) Hopper that has torso mass of 9 units,and (b) Performance of EACC-RARL, ESC-RARL and ERARL on target task with torso mass 9. }
	\label{fig:hoppermass8iter10}
\end{figure}

Regarding the parametric form of target tasks, the performance of the policies residing in the policy buffer can be anticipated. Coinciding with the performances seen in Figures \ref{fig:hoppermassiter10}(a), \ref{fig:hoppermassiter10}(f), \ref{fig:hoppermassiter10}(g), \ref{fig:hoppermass8iter10}(b), it is anticipated that the earlier policy iterations trained with less samples perform better as the distance between the target task and the source task increases in the task parameter space. We reproduced the original RARL experiment in \cite{pinto2017robust} where results are consistent with the performance of the last policy iteration trained with RARL (Figures \ref{fig:hoppermassiter10}(b), \ref{fig:hoppermassiter10}(c), \ref{fig:hoppermassiter10}(d)).

\begin{figure}[!htbp]
	\centering
	\subfigure[]{
		\includegraphics[width=.480\textwidth]{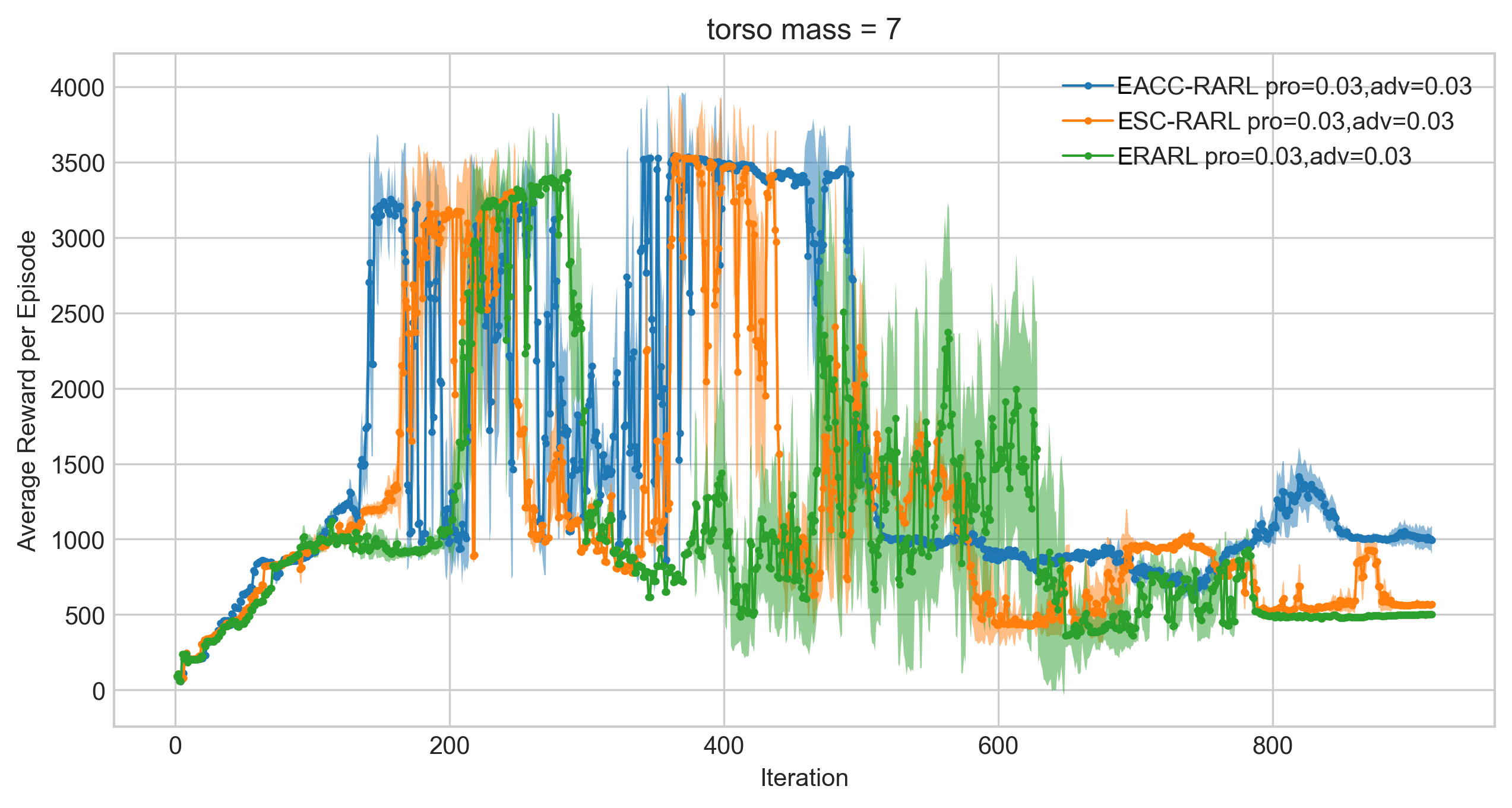}
	}
	\subfigure[]{
		\includegraphics[width=.480\textwidth]{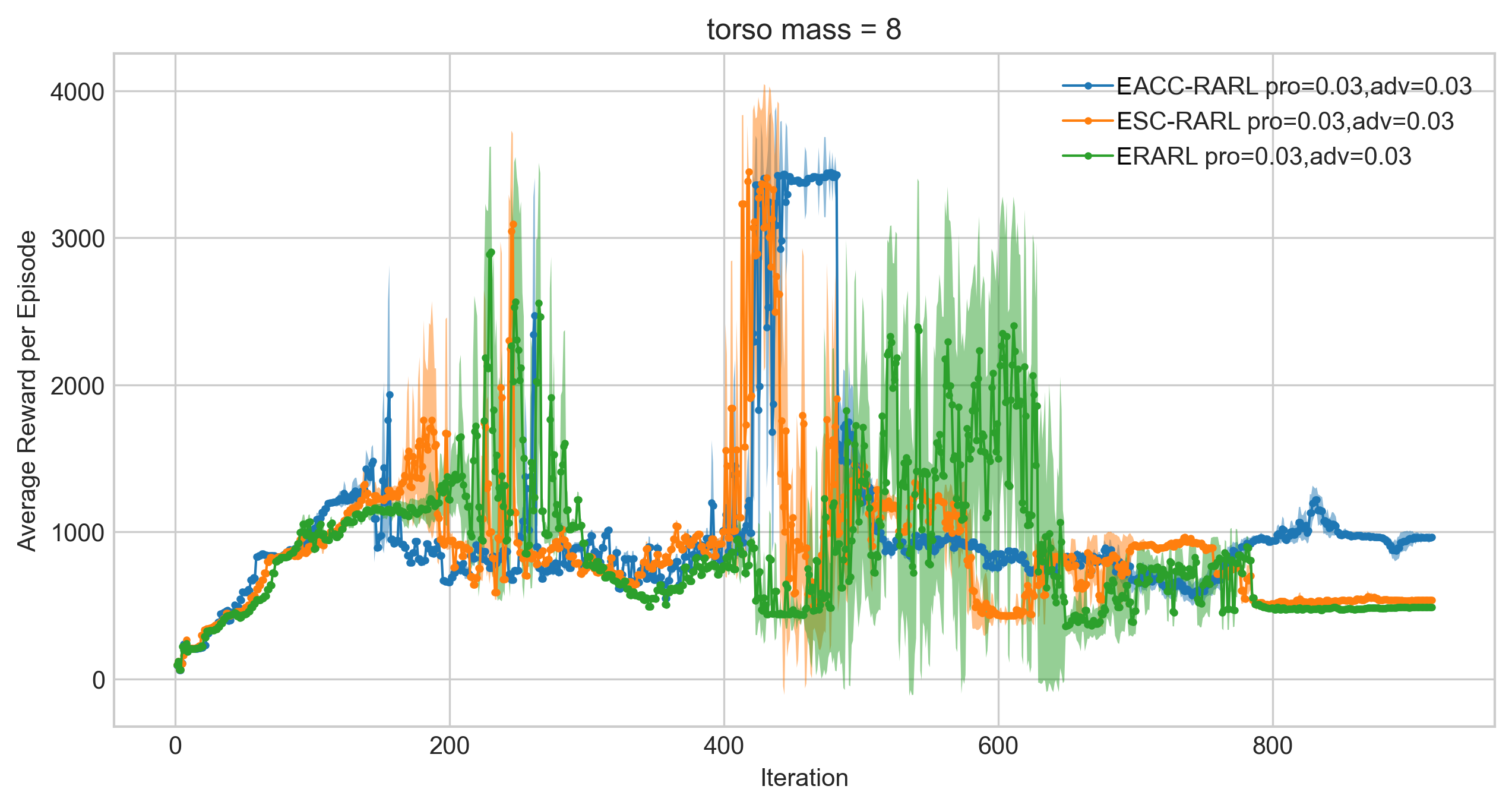}
	}
	\caption{Performance of EACC-RARL, ESC-RARL and ERARL on target tasks with torso mass (a)7 and (b)8}
	\label{fig:hoppermass67iter10}
\end{figure}

When the torso mass is increased to 6, significant target performance drop (Figure \ref{fig:hoppermassiter10}(f)) occurs until the last policy iteration where all policies are affected. This implies overfitting occurs and all policy iterations trained with SC-PPO can still perform optimally for policy iterations around 395. The best performing policy iterations of adversarial algorithms start to get accumulated in the range $[150,300]$ when torso mass is greater than 5 thus a mapping between the target task parameters and the policy iterations is highly probable.

In a harder environment with a torso mass of 7, the earlier iterations of the policy trained with ACC-RARL performs the best. Figure \ref{fig:hoppermassiter10}(g) shows that the range of the best-performing policy iterations is contracted more. 

\begin{table}[thbp]
	\vskip\baselineskip 
	\caption[Performance of the policy trained with ACC-RARL]{Performance of ACC-RARL}
	\begin{center}
		\begin{tabular}{|c|c|c|}\hline
			\textbf{Unit Mass}& \textbf{Iteration}& \textbf{Average Reward per Episode}\\\hline
			1 & \multirow{7}{*}{175} &$2921 \pm 12$\\  
			2 & &$3072 \pm 6$\\
			3 & &$3196 \pm 6$\\
			4 & & $3253 \pm 4$\\
			5 & &$3279 \pm 3$\\
			6 & & $3259  \pm 221$\\
			7 & & $2792 \pm 768 $\\\hline
		\end{tabular}
		\label{table:hopperacrarl}
	\end{center}
\end{table} 

Using  ME-RARL algorithms (EACC-RARL, ESC-RARL, and ERARL) increased the fluctuation of the average rewards for all algorithms as provided in Figure \ref{fig:hoppermass67iter10}. Because entropy bonus encourages exploration, the adversary takes more randomized actions which often change the protagonist's hopping behaviour. None of the adversarial techniques were successful in completing target tasks without entropy regularization where the torso mass of the hopper is increased to 8 and 9 units. EACC-RARL shows the highest performance on the 9 unit torso mass target task (Figure \ref{fig:hoppermass8iter10}(b)). The average reward per episode of the policy iterations trained with EACC-RARL is provided in Table \ref{table:hoppereacrarl}. These results show that earlier policy iterations show better performance as the target task becomes more challenging. Any target task from the set can be used as a proxy validation task for the tasks that belong to the same set.
\begin{table}[thbp]
	\vskip\baselineskip 
	\caption[Average Reward per Episode of the policy trained with ACC-RARL]{Performance of EACC-RARL}
	\begin{center}
		\begin{tabular}{|c|c|c|c|}\hline
			\textbf{Set}&	\textbf{Unit Mass}& \textbf{Iteration}& \textbf{Average Reward per Episode}\\\hline
			\multirow{4}{*}{1} &1 &\multirow{4}{*}{508} & $2872 \pm 36$\\
			  &2&  & $3377 \pm 501$\\
			 &3&  & $3196 \pm 13$\\
			 &4& & $3474 \pm 7$\\\hline
			 \multirow{3}{*}{2}&5 & \multirow{3}{*}{479}& $2704 \pm 764$\\
			 &6 & & $2764  \pm 901$\\
			 &7 & & $3425 \pm 3$\\\hline
			\multirow{2}{*}{3}&8 & \multirow{2}{*}{463}& $3423  \pm 5$\\
			 &9 & & $3283  \pm 500$\\\hline
		\end{tabular}
		\label{table:hoppereacrarl}
	\end{center}
\end{table}

First, using SC-PPO and early stopping we formulate a more competitive baseline for the adversarial algorithms. Then, we show the performance of one policy iteration ($175^{th}$) with high generalization capacity trained with the ACC-RARL algorithm in Table \ref{table:hopperacrarl} to demonstrate that a policy is capable of performing forward locomotion when torso mass is in the range of 1-7. Finally, using EACC-RARL and early stopping we have increased the target task success range used in RARL\cite{pinto2017robust} from $[2.5,4.75]$ to $[1,9]$. 

%% file: files/grav_hopper.tex
\begin{figure}[!htbp]
	\centering
	\subfigure[]{
		\includegraphics[width=.480\textwidth]{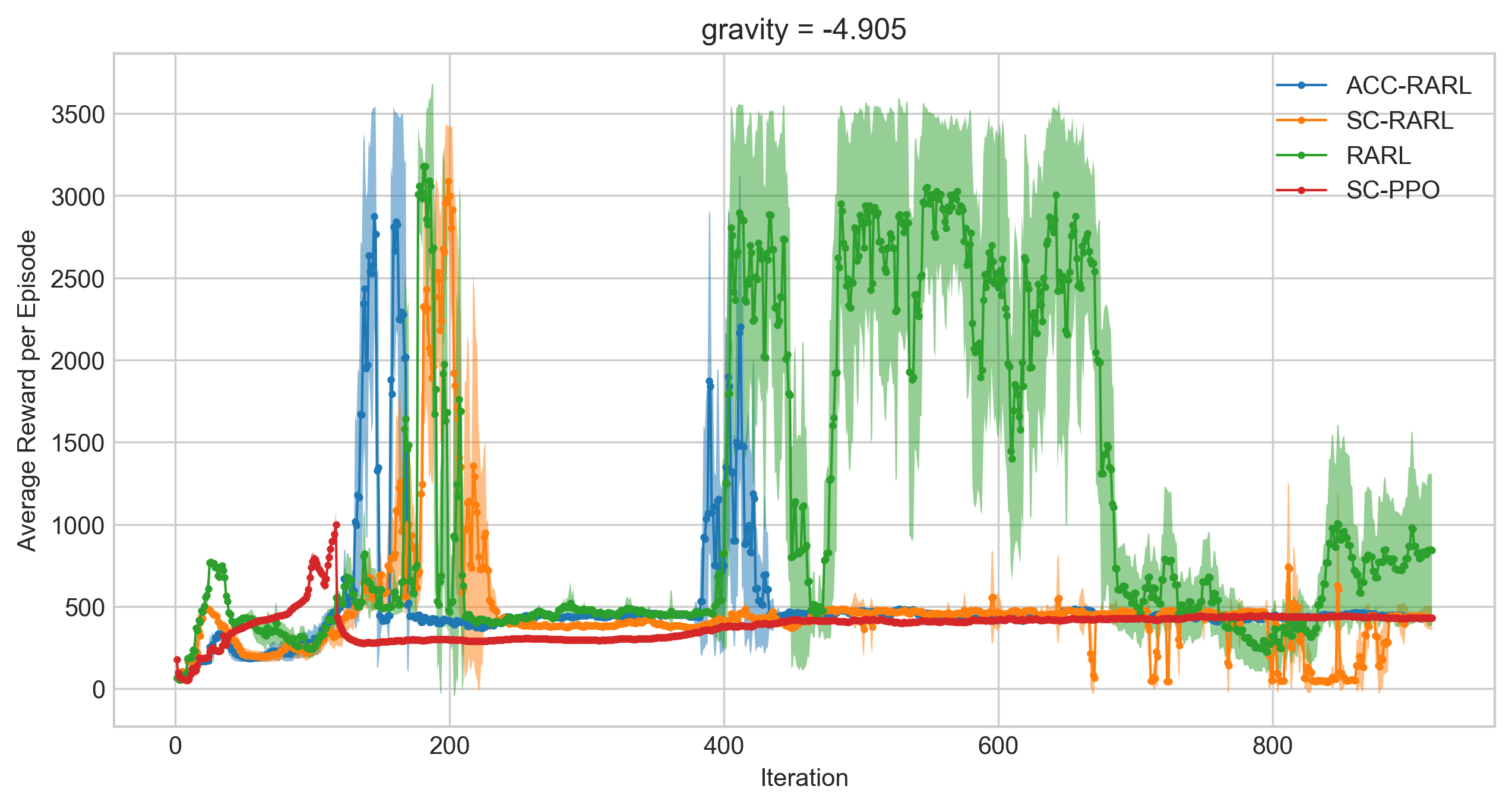}
	}
	\subfigure[]{
		\includegraphics[width=.480\textwidth]{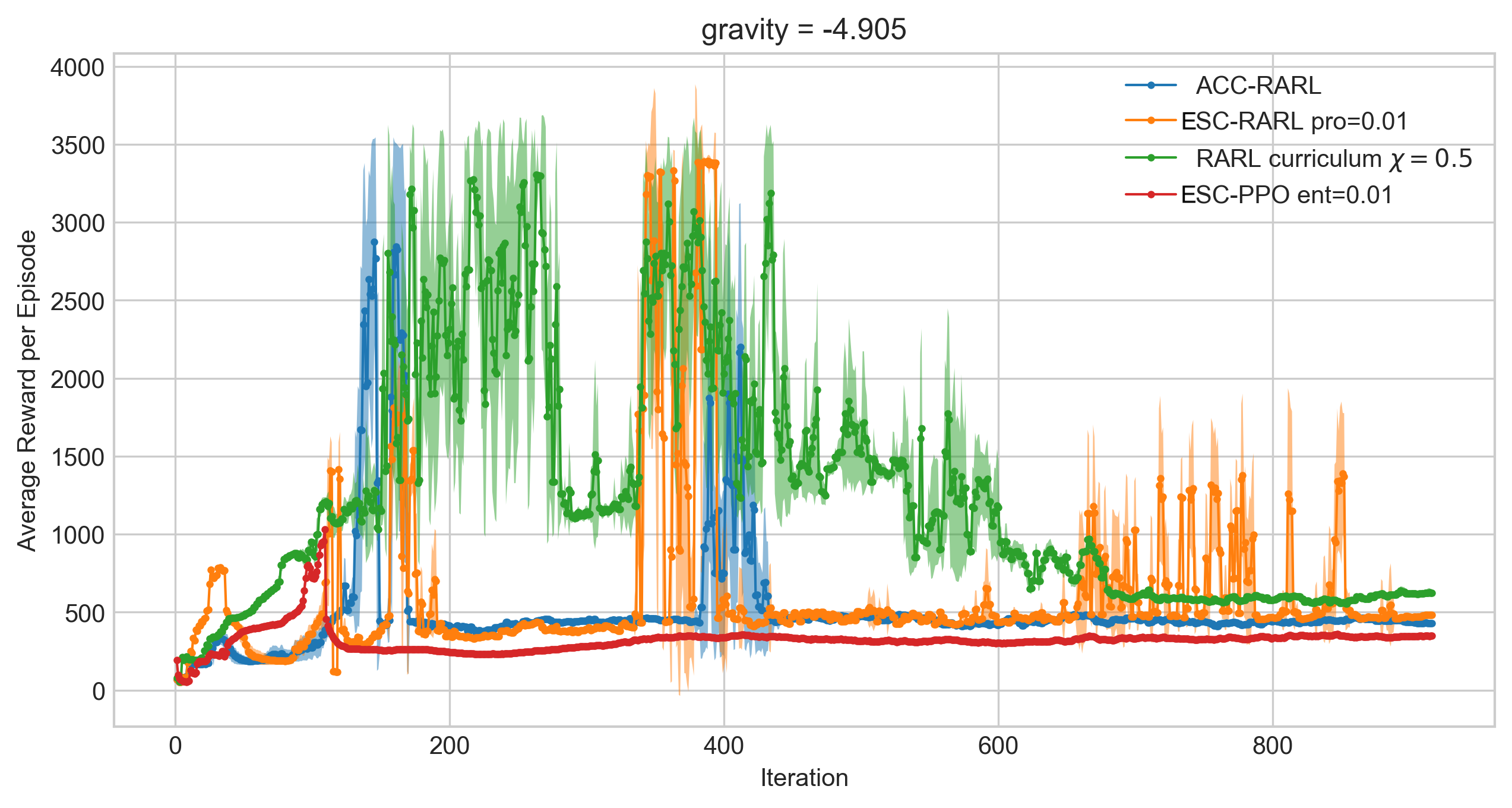}
	}
	\caption{Performance of (a)SC-PPO, ACC-RARL, SC-RARL and RARL, and (b)ESC-PPO, ACC-RARL, ESC-RARL and RARL with curriculum on target environment with gravity= -4.905 ($0.5G_{Earth}$)}
	\label{fig:10hoppergravity4905}
\end{figure}
\begin{figure}[!htbp]
	\centering
	\subfigure[]{
		\includegraphics[width=.480\textwidth]{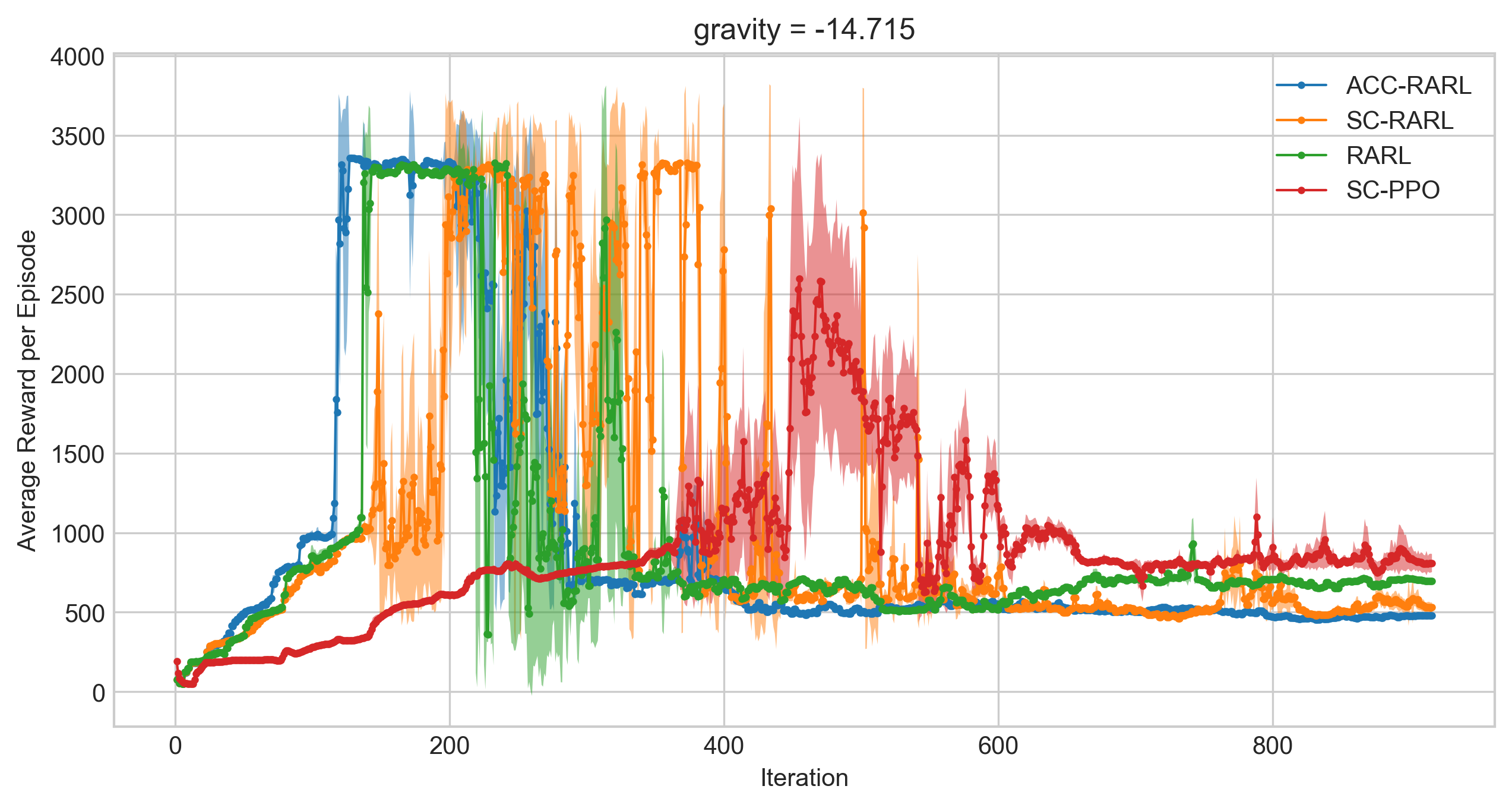}
	}
	\subfigure[]{
		\includegraphics[width=.480\textwidth]{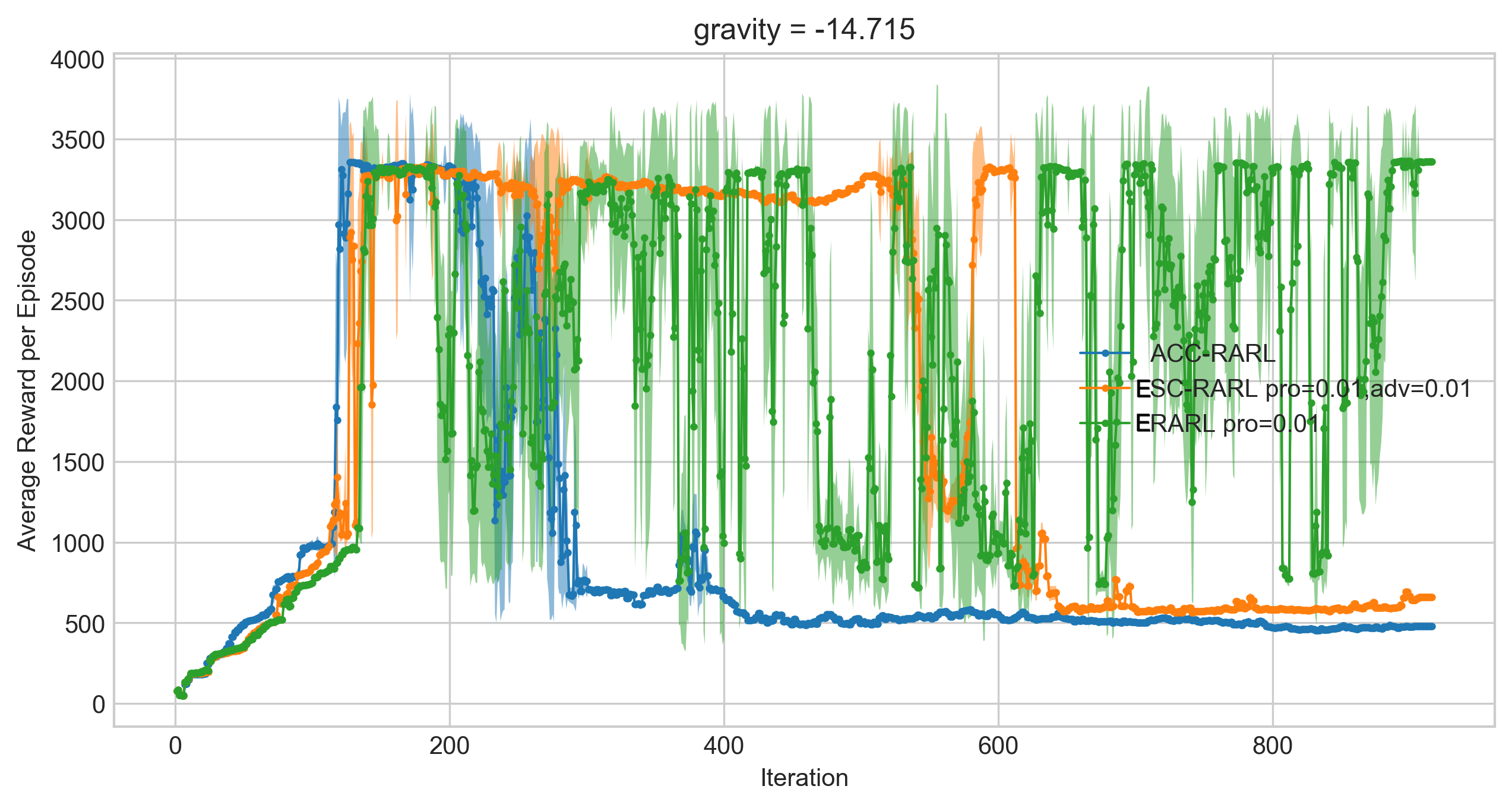}
	}
	\caption{Performance of (a)SC-PPO, ACC-RARL, SC-RARL and RARL, and (b) ACC-RARL, ESC-RARL and ERARL on target environment with gravity= -14.715 ($1.5G_{Earth}$)}
	\label{fig:10hoppergravity14715}
\end{figure}
In this set of experiments, we will assess our methods in a large range $[0.5G_{Earth},1.75G_{Earth}]$ of gravity tasks. In Learning Joint Reward Policy Options using Generative Adversarial Inverse Reinforcement Learning (OptionGAN) \cite{henderson2018optiongan}, the gravity of the generated environments is in the range of $0.5G_{Earth}-1.5G_{Earth}$ for both humanoid and hopper. The policy over options converges to 2 different policies for Hopper tasks in \cite{henderson2018optiongan}: one for lower and one for higher than the gravity of Earth indicating that the tasks are complex enough to be solved with different policies. For these tasks, we will use the same policy buffer created using SC-PPO and different variations of RARL for the hopper morphology experiments. 

The performance evaluations in Figures \ref{fig:10hoppergravity4905}(a) and \ref{fig:10hoppergravity4905}(b) prove that training with curriculum and maximum entropy changes the number of generalizable policy iterations in $0.5G_{Earth}$ target task. ESC-PPO included in Figure \ref{fig:10hoppergravity4905}(b) still performs poorly compared to adversarial techniques. Although the improvement is negligible, only for this case the entropy regularized SC-PPO (ESC-PPO) performs better than the SC-PPO.

Policies trained with SC-PPO and adversarial methods can be transferred to the target task with gravity= -14.715 in Figure \ref{fig:10hoppergravity14715}(a). Figure \ref{fig:10hoppergravity14715}b shows that using entropy regularized methods ESC-RARL and ERARL, not only increased the average reward per episode but also increased the number of generalizable policy iterations. 

\begin{figure}[!htbp]
	\centering
	\subfigure[]{
		\includegraphics[width=.480\textwidth]{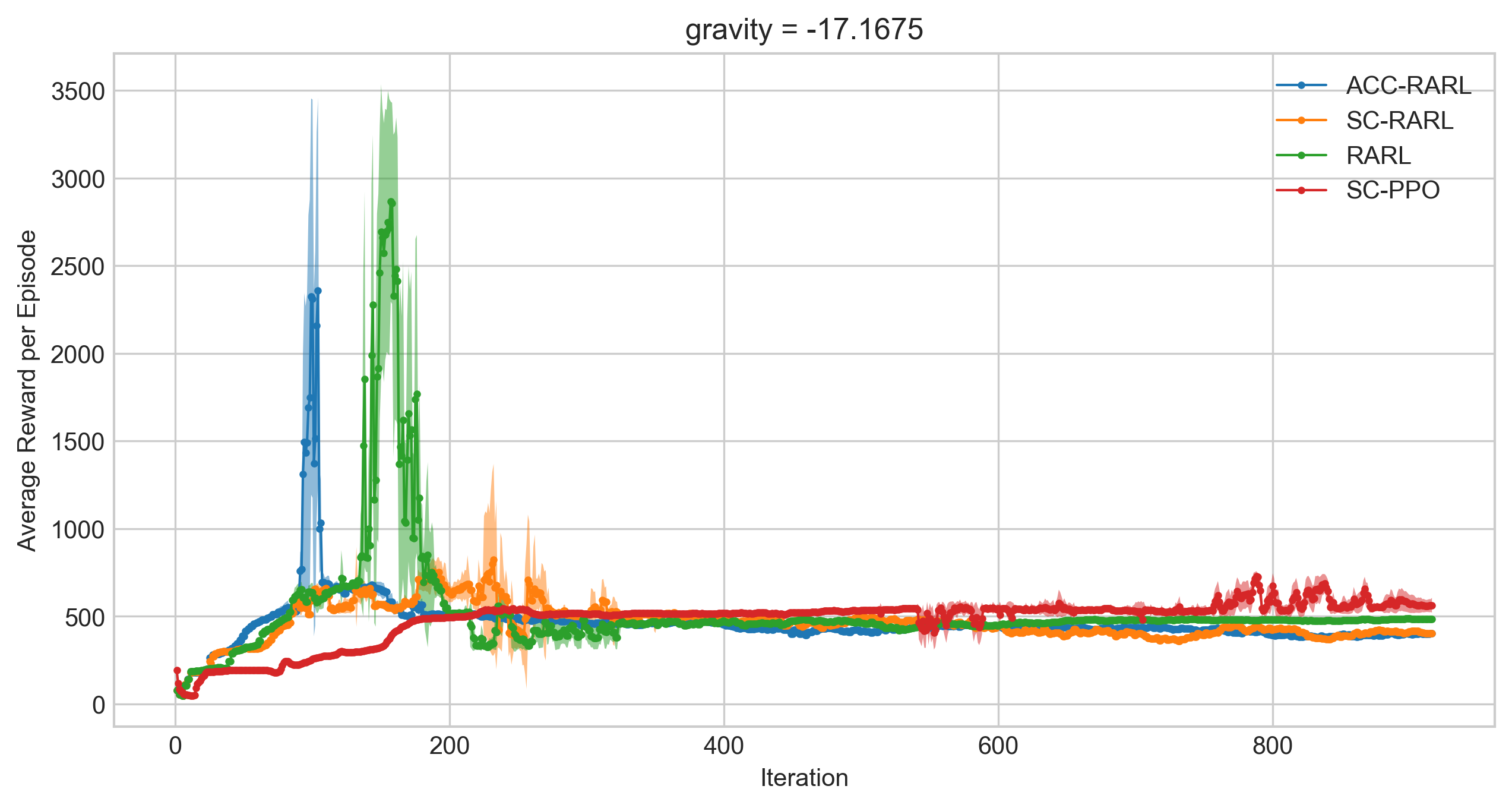}
	}
	\subfigure[]{
		\includegraphics[width=.480\textwidth]{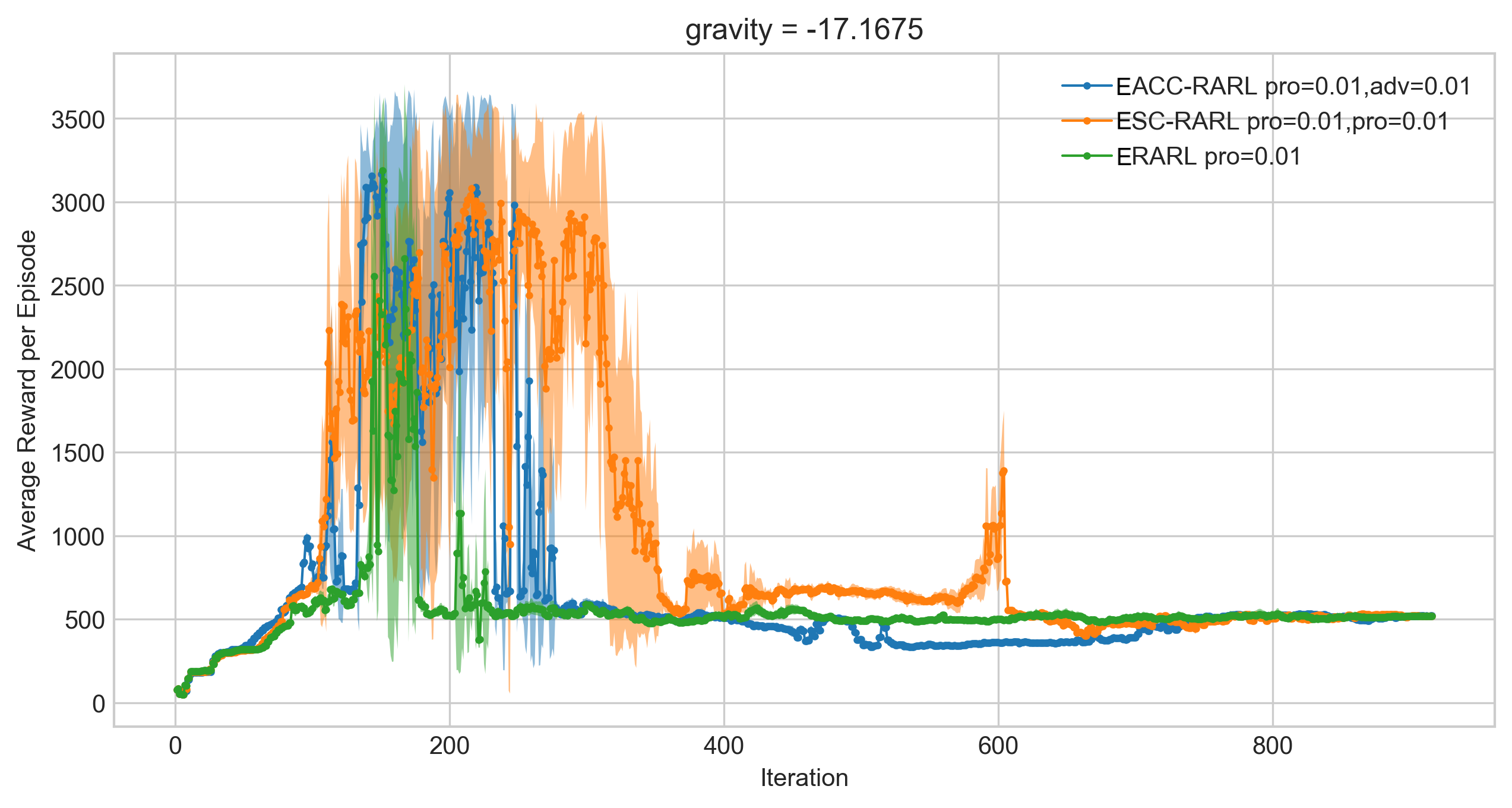}
	}
		\caption{Performance of (a)SC-PPO, ACC-RARL, SC-RARL and RARL, and (b) EACC-RARL, ESC-RARL and ERARL on target environment with gravity= -17.1675 ($1.75G_{Earth}$)}
	\label{fig:10hoppergravity171675}
\end{figure}

As the target task gets harder by moving further away from the source task, the best performing policy iterations are aggregated around earlier iterations similar to the torso mass and the delivery humanoid target tasks. This concavity is analogous to the convexity of the test error curve in supervised learning problems where earlier training iterations lead to underfitting and the later iterations overfit to the training set. The regularization effect of early stopping is pivotal in increasing the generalization capacity. The domain randomization in SC-RARL does not suffice for generalization in harder target tasks when gravity is $1.75G_{Earth}$ (Figure \ref{fig:10hoppergravity171675}(a)). We observe that encouraging the exploration of the protagonist policy and the adversary through the inclusion of entropy bonus increases the performance of adversarial algorithms in Figure \ref{fig:10hoppergravity171675}(b). Results in the gravity environments are in line with the morphology experiments. Above all, entropy regularized adversarial techniques, early stopping and strict clipping produce competitive results in hard target tasks.

%% file: files/conclusion.tex
In this study, we propose a set of methods to extract generalizable knowledge from a single source task. Training independently for each task is a customary sample inefficient process in deep RL. With each environment interaction, the agent's strategy of solving the source task is expected to advance. However, we find that the transferability of these strategies to similar tasks relies significantly on the number of environment interactions in the source task. The best strategy to solve the source task fails substantially in the target task.

Our experiments show that the performance of transfer RL algorithms is dramatically dependant on the choice of hyperparameters and the number of policy iterations used in the training of the source task. This dependency affects the reproducibility and evaluation accuracy of the algorithms in transfer RL. We addressed this issue in 19 different transfer RL experiments to show how experimental results can be manipulated in favor of an algorithm in the transfer RL domain. 

In our work, we proposed keeping a policy buffer to capture different skills because source task performance is not indicative of target task performance. Accordingly, transferring the policy with the best source task performance to the target task becomes a less adequate evaluation technique as the difference between the source and target task increases. In line with this, we suggest using a proxy validation task to extract the generalizable policies. To account for the early stopping regularization technique, we propose the inclusion of the policy iteration to the hyperparameter set. After this inclusion, we have retrieved the generalizable skills that generate high rewards in the target tasks. 

We introduced SC-PPO and early stopping as a regularization technique in transfer RL. Training SC-PPO promote the elimination of samples that overfit the source task. We experimentally show that these robust policies generate a higher jumpstart performance than the baseline in the target tasks. Using SC-PPO and early stopping, we transferred the forward locomotion skills of a standard humanoid to humanoids with different morphologies (a taller humanoid, a shorter humanoid, and a delivery humanoid), humanoids in environments with different gravities (0.5G, 1.5G, and 1.75G) and a humanoid in an environment with 3.5 times the ground friction coefficient of the source task. Moreover, our results show that we can increase the extrapolation range of Hopper morphology tasks from the range of [2.5,4.75] used in RARL to [1,6] via SC-PPO and early stopping. We applied the same technique to transfer the skills of a hopper to the target task with gravity 1.5G.

We conducted a comparative analysis with RARL with different critic methods, entropy bonus, and curriculum learning. Our results show that the choice of hyperparameters and training iteration affect the generalization capacity substantially in adversarial RL. Using entropy regularized ACC-RARL and early stopping via policy buffer, we have increased the extrapolation range of Hopper torso mass tasks from the range of [2.5,4.75] used in RARL to [1,9]. We prove that a hopper is capable of performing forward locomotion in a target task in the range of [0.5G,1.75G] by learning in the source task with regularized adversarial RL. Using entropy regularized adversarial RL, significantly increased the performance on target tasks and the number of generalizable policies extracted from the source task. 

We believe that the first step of determining the most promising policy parameters lies in the accurate parameterization of the source and target task space. In the future, we plan to investigate the relationship between the environment parameters and the source task training hyperparameters.